\pdfoutput=1
\documentclass[11pt]{article}

\usepackage{ACL2023}

\usepackage{times}
\usepackage{latexsym}

\usepackage[T1]{fontenc}
\usepackage[utf8]{inputenc}

\usepackage{microtype}

\usepackage{inconsolata}

\usepackage{amsmath}

\usepackage{amssymb}
\usepackage{pifont}
\newcommand{\xmark}{\ding{55}}

\usepackage{subcaption}
\usepackage{booktabs}
\usepackage{xspace}
\usepackage{graphicx}
\usepackage{multirow}
\usepackage{adjustbox}
\usepackage[skins]{tcolorbox}
\usepackage{xcolor}
\usepackage{algpseudocode}
\usepackage[ruled,vlined,linesnumbered]{algorithm2e}
\usepackage{mathdots}

\newcommand{\gptthreepointfive}{\textsc{GPT-3.5}\xspace}
\newcommand{\gptfouromini}{\textsc{GPT-4o-mini}\xspace}
\newcommand{\llamathree}{\textsc{Llama3-8B}\xspace}
\newcommand{\llamathreepointone}{\textsc{Llama3.1-8B}\xspace}
\newcommand{\llamathreepointtwo}{\textsc{Llama3.2-11B}\xspace}
\newcommand{\gemmatwo}{\textsc{Gemma2-9B}\xspace}
\newcommand{\method}{Character-Level Manipulation via Divide and Conquer\xspace}

\newcommand{\deltask}{\texttt{Deletion}\xspace}
\newcommand{\instask}{\texttt{Insertion}\xspace}
\newcommand{\subtask}{\texttt{Substitution}\xspace}
\title{Enhancing LLM Character-Level Manipulation via Divide and Conquer}

\author{Zhen Xiong$^\dagger$ \ \ \ \ Yujun Cai$^\ddagger$ \ \ \ \ Bryan Hooi$^\|$ \ \ \ \ Nanyun Peng$^\mathsection$ \\ \textbf{Zhecheng Li}$^\delta$ \ \ \ \ \textbf{Yiwei Wang}$^\mathparagraph$ \\
$^\dagger$ University of Southern California \quad $^\ddagger$ The University of Queensland \\ $^\|$ National University of Singapore \quad $^\mathsection$ University of California, Los Angeles \\ $^\delta$ University of California, San Diego \quad $^\mathparagraph$ University of California, Merced \\
  \texttt{xiongzhe@usc.edu} \\}

\begin{document}

\maketitle

\begin{abstract}

Large Language Models (LLMs) have demonstrated strong generalization capabilities across a wide range of natural language processing (NLP) tasks. However, they exhibit notable weaknesses in character-level string manipulation, struggling with fundamental operations such as character deletion, insertion, and substitution. These challenges stem primarily from tokenization constraints, despite the critical role of such operations in data preprocessing and code generation. Through systematic analysis, we derive two key insights: (1) LLMs face significant difficulties in leveraging intrinsic token knowledge for character-level reasoning, and (2) atomized word structures can substantially enhance LLMs' ability to process token-level structural information. Building on these insights, we propose \method, a novel approach designed to bridge the gap between token-level processing and character-level manipulation. Our method decomposes complex operations into explicit character-level subtasks coupled with controlled token reconstruction phases, leading to significant improvements in accuracy. Without additional training, our method significantly improves accuracies on the \deltask, \instask, and \subtask tasks. To support further research, we open-source our implementation and benchmarks.%\footnote{\url{https://github.com/Eric2i/CharDC}}

\end{abstract}

\section{Introduction}

Recent years have witnessed remarkable success of Large Language Models (LLMs) across diverse NLP tasks~\cite{language2020, emergent2022, finetuned2022}. However, these models show surprising weaknesses in seemingly simple character-level manipulation tasks. For example, when prompting models to insert `a' after every `e' in the word "intelligence", even one of the state-of-the-art LLMs, ChatGPT-4o, returns a wrong answer: "intellaigenca". Such failures are not isolated cases but reflect a systematic limitation in LLMs' ability to perform basic character-level operations on word.

Understanding this limitation requires examining how LLMs process text in details. Modern LLMs convert input text into token sequences using tokenization algorithms, like commonly used BPE~\cite{neural2016} and SentencePiece~\cite{sentencepiece2018}, which optimize vocabulary coverage to some extent. For instance, "linguistics" might be first broken into "$\underline{\text{ling}}\cdot \underline{\text{u}} \cdot \underline{\text{istics}}$" and then tokenized into [3321, 84, 7592]\footnote{o200k\_base tokenizer}. While this approach effectively handles diverse vocabularies, it creates a fundamental barrier to character-level knowledge access. Interestingly, LLMs can exhibit high accuracy in spelling tasks, suggesting they learned some character-level knowledge through its training phases. Yet, according to our experiments, this capability rarely transfers to active character manipulation.

Character-level operations form the foundation of many text-processing systems. In software engineering, code generation and debugging often require precise character modifications for syntactic and semantic auto-correction~\cite{codex2021}. Data pre-processing pipelines rely on character manipulation for text normalization and cleaning~\cite{jellyfish2024}. Educational applications need character-level editing for spelling completion and grammar error correction~\cite{gector2020, application2023}. Despite the prevalence of these applications, existing research has primarily focused on assessing LLMs' Character-level operations performance through benchmarks, like CUTE~\cite{cute2024} for instance, without deeply investigating the underlying mechanisms or proposing effective solutions.

\begin{figure}[!t]
    \centering
    \includegraphics[width=1\linewidth]{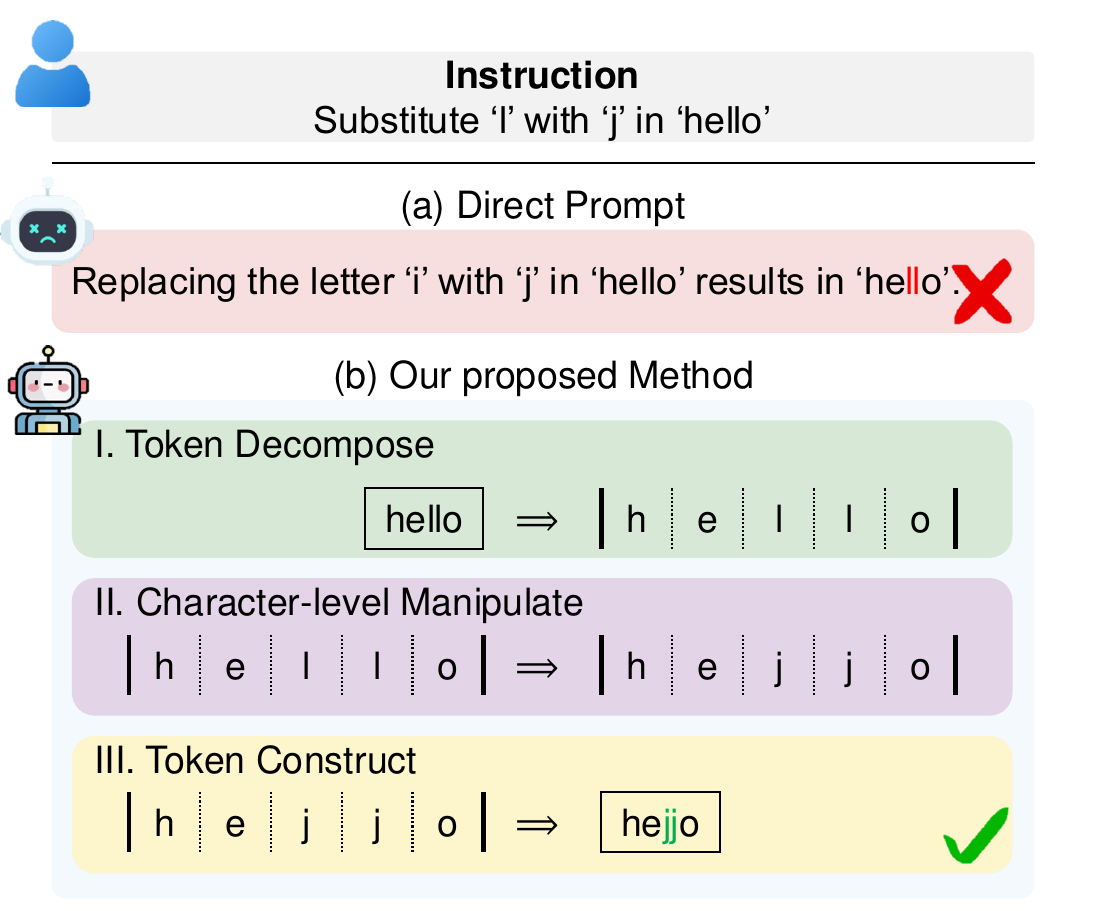}
    \caption{ (\textit{Upper}) Given a character-level token manipulation instruction,  (\textit{Lower}-(a)) direct prompting fails with an incorrect answer while our (\textit{Lower}-(b)) proposed Token Character-Aware Decomposition (ToCAD) method is more robust to such token manipulation tasks.}
    \label{fig:1}
\end{figure}

Our systematic analysis reveals that LLMs consistently perform well in character-by-character spelling tasks, and that atomized word forms can enhance their reasoning in character-level tasks. Building on these insights, we propose \method, a novel approach designed to bridge the gap between token-level knowledge and character-level manipulation. Specifically, our method decomposes complex operations into three carefully designed stages: token decomposition, character-level manipulation, and token reconstruction. This method effectively integrates token-level processing with character-level operations without requiring additional training. See Figure~\ref{fig:1} for an illustrative example.

Comprehensive experiments validate the effectiveness of our approach. With GPT-3.5, our method significantly outperforms existing methods across \deltask, \instask, and \subtask tasks. Our analysis further reveals specific error patterns that provide valuable insights into LLMs' character-level processing mechanisms.

\begin{itemize}
    \item We present the first systematic analysis of character manipulation challenges in LLMs and identify a specific structural variation form that boosts LLM's reasoning ability on character-level knowledge.
    \item We propose a novel zero-shot approach that significantly improves character operation capabilities while being compatible with existing models without extra finetuning.
    \item We provide extensive experimental analysis that not only validates our method but also offers insights for future research in enhancing LLM character-level reasoning. 
\end{itemize}

\section{Related Works}

With the rise of "seq2seq"~\cite{seq2014} models built on the Transformer~\cite{attention2017} architecture, natural language processing has stepped into its new the era of Large Language Model (LLM). Instruction fine-tuning technique~\cite{language2020} has enabled LLMs to demonstrated remarkable generality: they show the potential to outperform human performance in tasks such as mathematics~\cite{math2024}, programming~\cite{programming2023}, and logical reasoning~\cite{logicbench2024}. However, LLMs can still make naive mistakes on simple problems by generating seemingly correct but nonfactual or wrong answer~\cite{hallucination2024}.

Text preprocessing in modern language learning models (LLMs) primarily employs subword tokenization methods~\cite{google2016, subword2018}, with byte pair encoding (BPE)~\cite{neural2016} being one of the most widely used approaches. However, the subword tokenization paradigm has notable limitations that can hinder the nuanced understanding of internal structure of words~\cite{tokenization2024}. In this paper, we explore these limitations through a series of character-level tasks designed to assess LLMs' comprehension of words at the character level.

Current benchmarks that evaluate large language models' understanding of token composition reveal significant flaws and shortcomings in LLMs that use subword units as tokens. For instance, LMentry~\cite{lmentry2022} tests whether LLMs can distinguish between the first and last letters of a word or generate words containing a specific letter. Meanwhile, CUTE~\cite{cute2024} introduces more challenging tasks, such as asking LLMs to replace, delete, insert, or swap letters within words. The results demonstrate that even the most advanced LLMs still have considerable room for improvement on non-trivial token benchmarks. Our paper goes beyond evaluation, presenting not only underlying mechanism analysis but also effective methods.

A significant amount of research has been conducted on character-level models, where each individual character is treated as a separate token. Although the character-level models exhibited potential for better generalization, especially in scenarios involving rare words, but they often struggled with efficiency and performance on tasks that require more abstract linguistic knowledge.

\section{Analysis}
\label{sec:ProblemAnalysis}

Large Language Models have demonstrated remarkable capabilities in complex NLP tasks, from reasoning to coding. However, a simple character manipulation task reveals their surprising limitations. When GPT-4o receives an instruction to insert `a' after every `e' in the word "intelligence", it comes up with a wrong answer, "intellaigenca". This intriguing phenomenon motivates us to systematically investigate why modern LLMs struggle with seemingly simple character manipulations despite their sophisticated abilities.

\subsection{Challenges in Character-Level Reasoning for LLMs}
\label{sec:WeakUnderstanding}

To understand LLMs' capabilities in handling characters within tokens, we first design diagnostic experiments using small open-sourced models as our primary study subject. The experiments aim to probe two aspects of character-level understanding: 1) the ability to spell out characters sequentially and 2) the ability to reason about individual characters within a word.

\paragraph{Spelling} In the spelling task, we evaluate the model's capability to decompose words into their constituent characters. \gemmatwo achieves an impressive 97.4\% accuracy on 814 single token English word, suggesting a strong ability to serialize words into character sequences. This high performance leads to an intuitive assumption that the model possesses a robust understanding of character-level composition.

\begin{figure*}[!t]
    \centering
    \includegraphics[width=\textwidth]{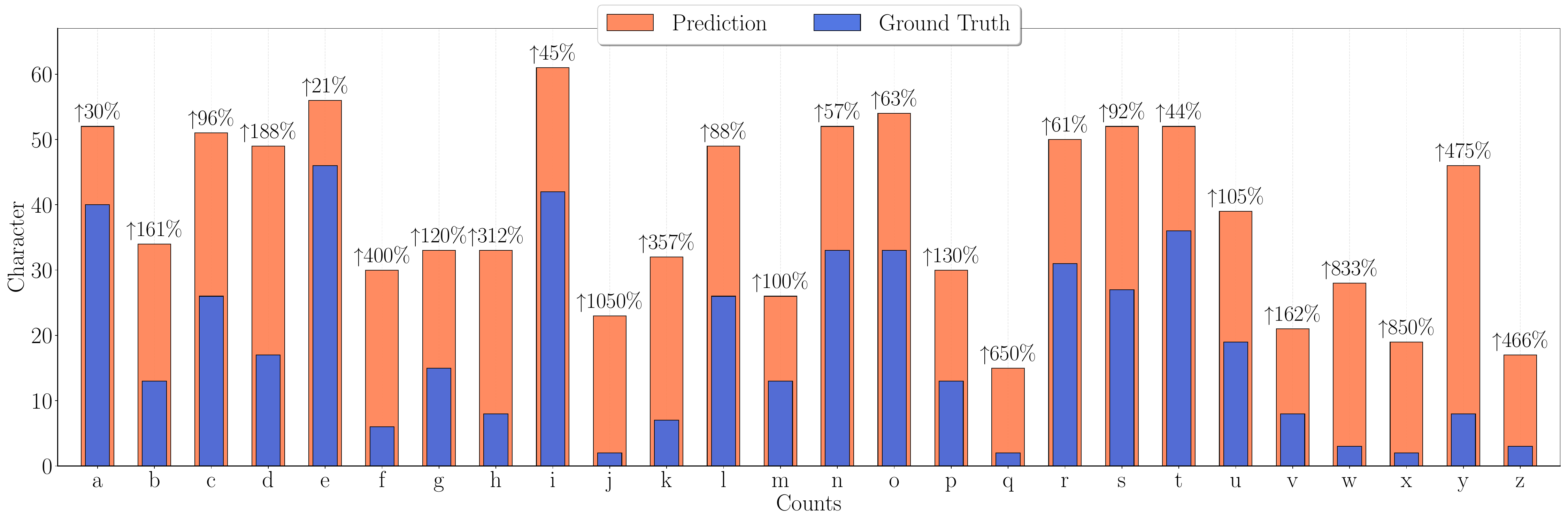}
    \caption{A more comprehensive character retrieval experiment conducted on \gemmatwo. The model tends to mistakenly identify characters that are not present in a word as being part of it. The percentages represent the ratio of false positives to true positives for each letter.}
    \label{fig:counting_bar}
\end{figure*}

\paragraph{Reasoning} However, this assumption quickly breaks down when we examine the model's ability to verify if a certain letter exists in a word, see Figure \ref{fig:char-verify} for an extreme example. Statically, the model frequently reports non-existent characters, leading to false positive rates up to 1050\% higher than true positives for certain characters, as shown in Figure \ref{fig:counting_bar}.

\begin{figure}[!t]
    \centering
    
\begin{tcolorbox}[width=\linewidth, fonttitle = \small\bfseries, title=Word Spelling,colframe=gray!2!black,colback=gray!2!white,boxrule=1pt,boxsep=0pt,left=5pt,right=5pt,fontupper=\footnotesize, halign title = flush center]
\textbf{Prompt:} Spell the word "hello".
\tcbline
\textbf{\color[RGB]{0,0,0}{Output}:} The word "hello" spelled out is: H-E-L-L-O{\color[RGB]{0,255,0}{\checkmark}}
\end{tcolorbox}

\begin{tcolorbox}[width=\linewidth, fonttitle = \small\bfseries, title=Character Retrieval,colframe=gray!2!black,colback=gray!2!white,boxrule=1pt,boxsep=0pt,left=5pt,right=5pt,fontupper=\footnotesize, halign title = flush center]
\textbf{Prompt:} Is the letter "\textit{a}" in "hello"?
\tcbline
\textbf{\color[RGB]{0,0,0}{Output}}: \textcolor{red}{Yes}, the letter 'a' is present in the word 'hello'. Let's think through it step by step: The word 'hello' consists of 5 letters: h, e, l, o, and l.
Check for the presence of the letter 'a': 
1) The first letter is 'h', which is \textcolor{green}{not} 'a'.
2) The second letter is 'e', which is \textcolor{green}{not} 'a'.
3) The third letter is 'l', which is \textcolor{green}{not} 'a'.
4) The fourth letter is 'l', which is \textcolor{green}{not} 'a'.
5) The fifth letter is 'o', which is \textcolor{green}{not} 'a'.
Conclusion: After checking each letter individually, we find that the letter 'a' is present as the \textcolor{red}{fifth} letter in the word 'hello'.
Therefore, the letter 'a' is \textcolor{red}{indeed present} in the word 'hello'.
\tcbline
\textbf{\color[RGB]{0,0,0}{Answer}} Yes {\color[RGB]{255,0,0}{\xmark}}
\end{tcolorbox}

\caption{\textit{(Upper)} Word spelling task performed with \llamathreepointone. \textit{(Bottom)}Single character retrieval experiment conducted on \llamathreepointone. Intriguingly, the model concludes with a wrong answer even if its reasoning process is correct. We also observed similar incorrect answer, without reasoning process though, from \gemmatwo, \llamathreepointtwo and \gptthreepointfive.}

\label{fig:char-verify}
\end{figure}

% \begin{figure}[!t]
%     \centering
%     \includegraphics[width=\linewidth]{fig/counting_hbars.pdf}
%     \caption{A more comprehensive character retrieval experiment conducted on \gemmatwo. The model tends to mistakenly identify characters that are not present in a word as being part of it. The percentages represent the ratio of false positives to true positives for each letter.}
%     \label{fig:counting_hbar}
% \end{figure}

\begin{figure}[!t]
    \centering
    \includegraphics[width=0.8\linewidth]{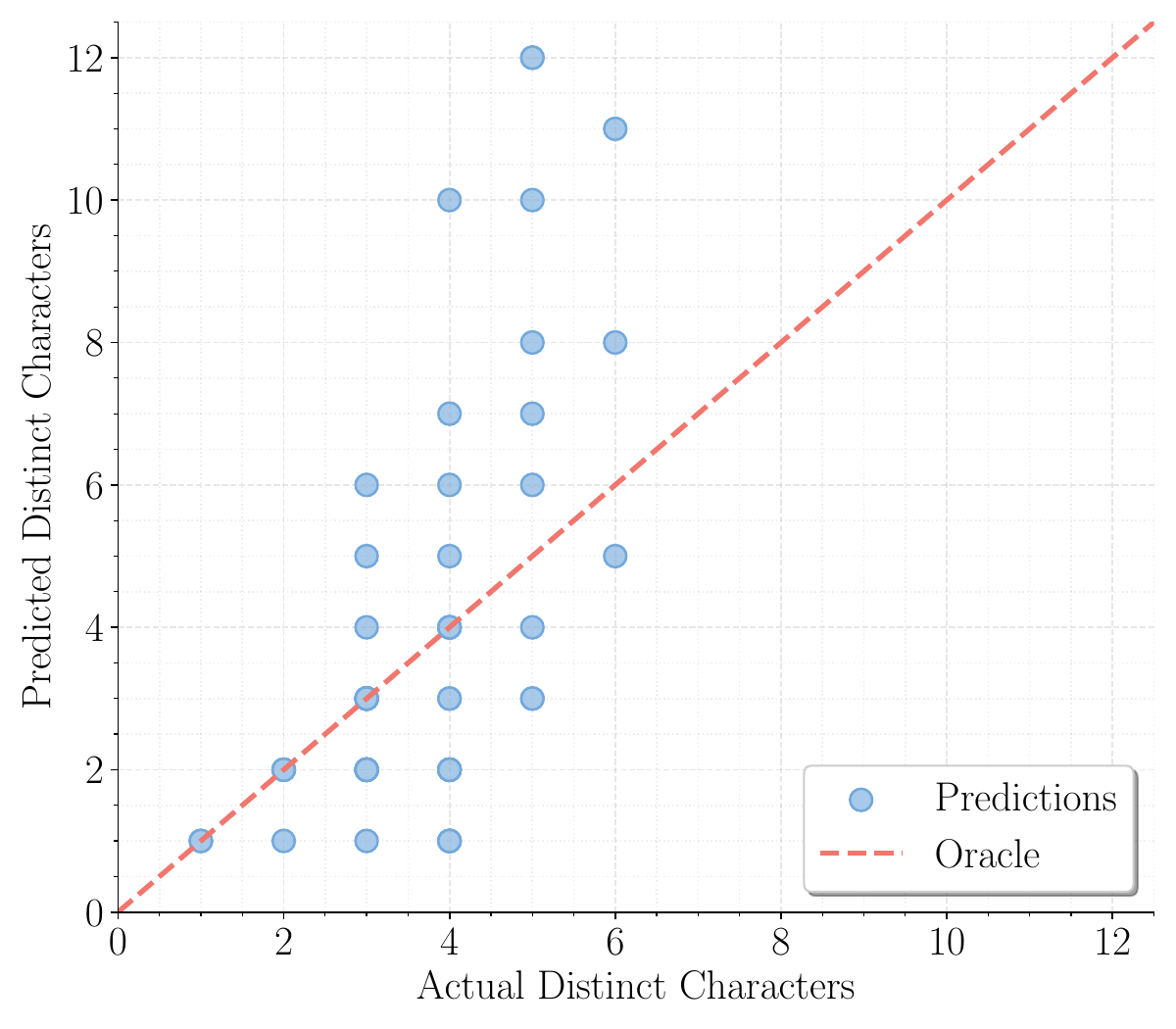}
    \caption{LLM (\gemmatwo) predicted distinct characters vs the actual number of distinct characters within a word. The scatter points illustrate the deviation of predictions from the reference line (Oracle), which increases w.r.t the token length. This demonstrates that the model's performance deteriorates as token length grows.}
    \label{fig:counting_calibration}
\end{figure}

Further investigation reveals a systematic error pattern: our analysis demonstrates that verification accuracy deteriorates significantly as token length increases. The relationship between token length and prediction accuracy, illustrated in Figure \ref{fig:counting_calibration}, suggests a fundamental limitation in how LLMs actively utilize character-level knowledge of tokens at different length.

\subsection{Atomized Word Structure Enhances Character-Level Reasoning for LLMs}

Having established that LLMs internally encode the compositional structure of words but do not actively leverage this information when processing related queries, we now turn to explore methods to activate this \textit{hidden} potential and unlock character-level reasoning capabilities.

To achieve this, we first investigate how LLMs comprehend a word’s internal structural information by systematically examining the impact of orthographic variations on the model’s internal lexical representation. Specifically, we design a controlled perturbation method that generates different segmentation patterns of the same word while preserving its character sequence.

% notation preparation
% Current Language Model (LLM) which follows the next-token prediction paradigm can be formally characterized as a conditional probability distribution over sequences of tokens. Given input tokens $x_{<t}=\{x_0, x_1, \cdots, x_{t-1}\}$ and a pre-trained LLM denoted as $p$, the probability distribution of the next token can be expressed as $p(x_t | x_{<t})$. Inside the decoder-only transformer-based language model, there are normally a embedding layer $E$ for token encoding, $N$ stack transformer layers for feature extraction and aggregation, and finally an FFN network to predict the next-token's distribution. We denote the embeddings for each token as $H_0 = \{h_{0}^{(0)}, h_{1}^{(0)}, \cdots, h_{t-1}^{(0)}\}$. For layer $j$, we denote the output of that layer as $H_j$ naturally.
% notation preparation

% We get started with the following question template: "Remove the letter `$L$' from the string '$W$'. Only return the result." In practice, the single letter $L$ in a randomly sampled letter from the word $W$. For each prompt instance with specific $W$, we only extract its last token's \footnote{due to subword tokenization algorithm, a word can be divided multiple subwords. We use $W_{-1}$ to denote the last subword token.} hidden states across every layers as its inner representation: $H_{W_{-1}} = \{h_{W_{-1}}^{(0)}, h_{W_{-1}}^{(1)}, \cdots, h_{W_{-1}}^{(t-1)}\}$.

For case-level comparison, we define a perturbation degree k that determines the percentage of adjacent character pairs to be separated by whitespace. Given a word with length L, we randomly select $\lfloor(L-1)k\%\rfloor$ pairs of adjacent characters to insert whitespace between them. For instance, given the word "information", different perturbation degrees yield:

\begin{itemize}
    \item 0\% perturbation:"information" (original form)
    \item 25\% perturbation: "in for mation" (2 spaces)
    \item 50\% perturbation: "in fo rm a tion" (4 spaces)
    \item 100\% perturbation: "i n f o r m a t i o n" (fully atomized form)
\end{itemize}

\begin{figure}[!tb]
    \centering
    \includegraphics[width=\linewidth]{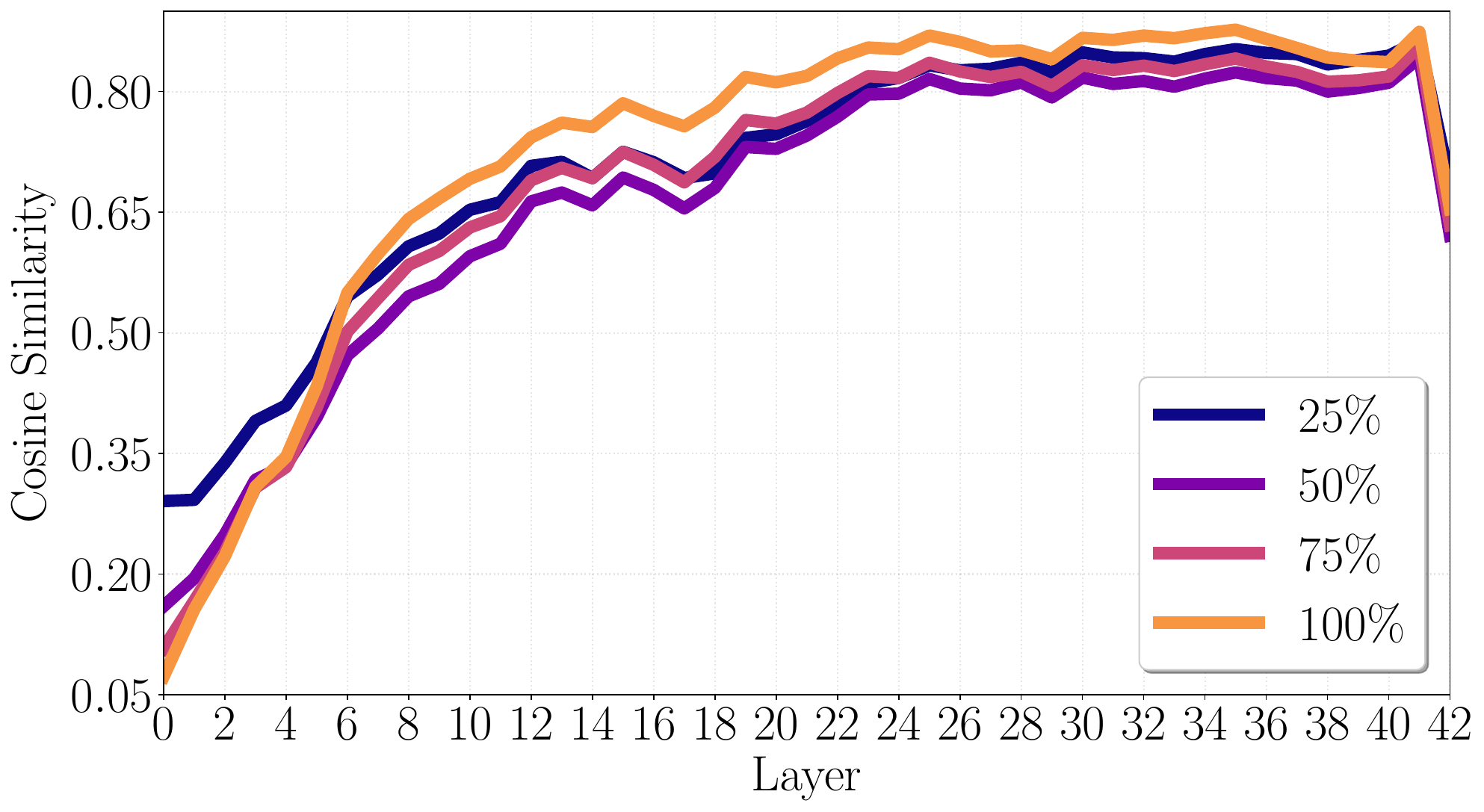}
    \caption{The cosine similarity of hidden states between the original word form and its orthographic perturbed forms at varying degree across different layers of \gemmatwo.}
    \label{fig:cos_sim}
\end{figure}

This perturbation framework allows us to systematically analyze how different segmentation patterns affect the model's internal representations. We examine the cosine similarities between the hidden states of perturbed versions and the original word across different layers of the model (see Figure~\ref{fig:cos_sim}). At early layers ($l \in [0, 4]$), the similarities are naturally low since we only extract the last token's representation. In middle layers ($l \in [5, 12]$), similarities increase dramatically due to word-level detokenization processes~\cite{lexicon2024}. Interestingly, in later layers ($l \geq 13$) when the lexicon representation stabilize, we observe that the fully atomized form (100\% perturbation) shows the strongest similarity to the original word. This suggests that the model maintains a strong internal connection between a word and its character-by-character spelling, aligning with our observations from Section~\ref{sec:WeakUnderstanding} about LLMs' high spelling accuracy.

\begin{figure*}[!tb]
    \centering
    \includegraphics[width=1\textwidth]{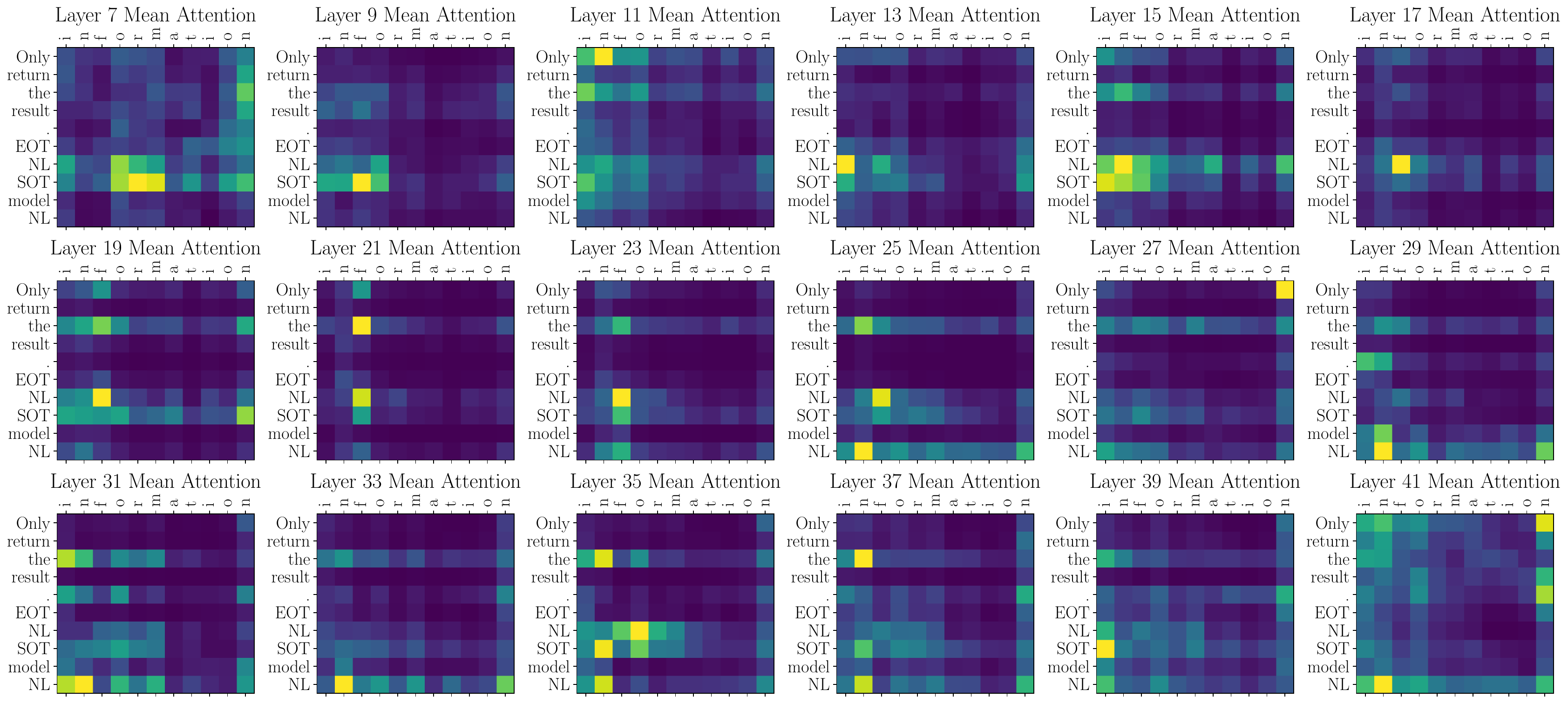}
    \caption{Averaged attention matrix across different layers with a special focus on tokens of interest. The darker the color of the heat map cell represents smaller the attention value, and vice versa. In our case, the token "f" in the x-axis is the target letter to remove from the word "information". Meanwhile, "in" is the first expected output token. See more examples in the appendix~\ref{apdix:attentions}.}
    \label{fig:attn_mat}
\end{figure*}

\begin{figure}[!tb]
    \centering
    \includegraphics[width=0.9\linewidth]{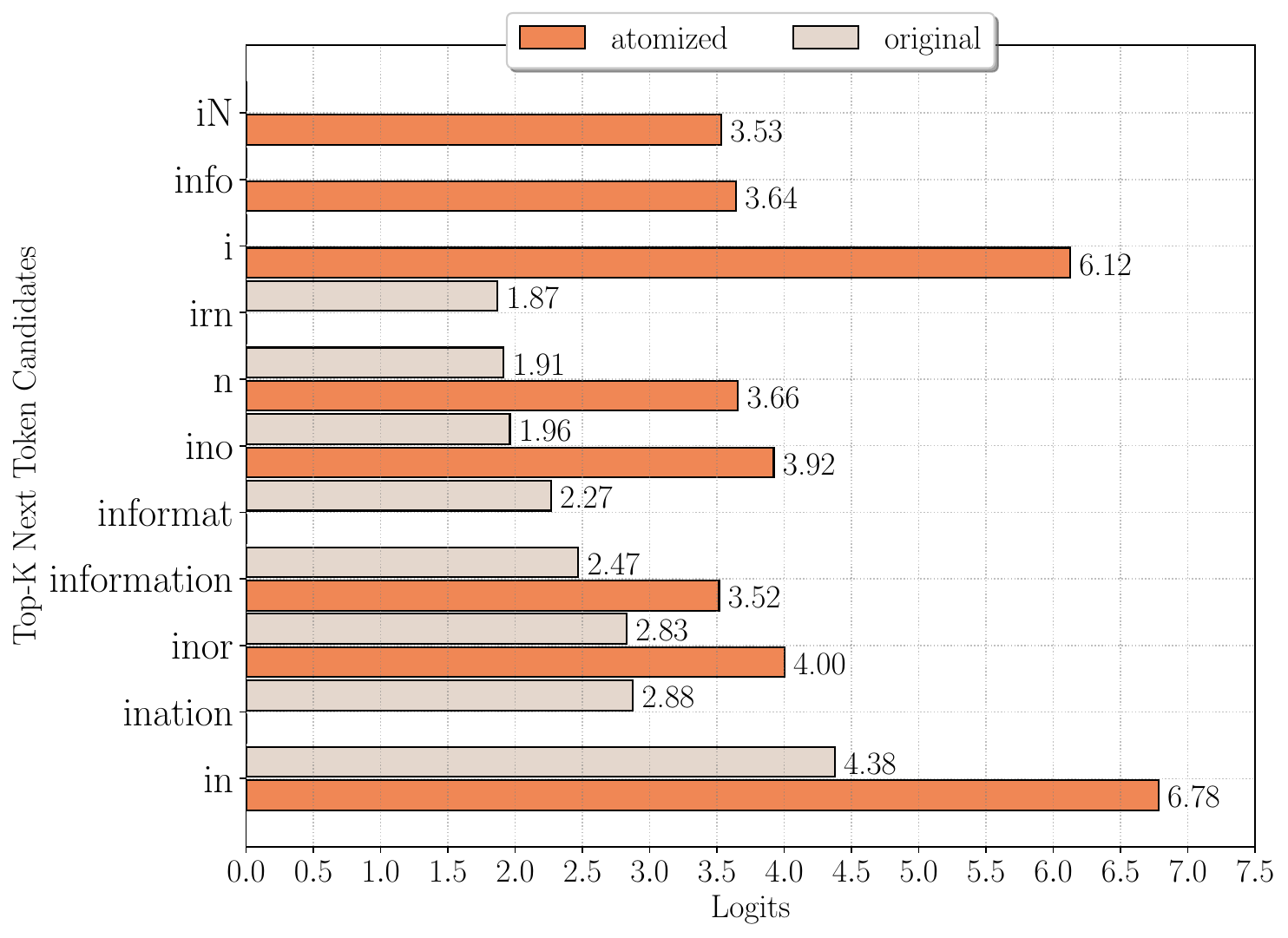}
    \caption{Logits for top-5 next token candidates using original word form v.s. atomized form. In our case, the sub-word "in" is the correct next-token prediction.}
    \label{fig:next_token_prob}
\end{figure}

With the special orthographical structure of the atomized word, we are now able to unveil the possible underlying process of how LLM deal with character-level knowledge reasoning, see Figure \ref{fig:attn_mat} as a qualitative analysis example. Based on the attention map across different layers, the model starts to shift its attention from the whole word to the specific character of interest, in our case, the letter to be removed. In later layers ($l > 25$), we observe that the last input token starting to pay more attentions to the token which is closely related to the ground truth. Finally, we also observed that with the atomized word, LLM did achieve better confidence, see Figure~\ref{fig:next_token_prob} for comparison of the probabilities of the top-5 output token candidates in our example.

Quantitatively, on larger scale experiments with 1K word samples, compared to the original word form, atomized words achieved a 143\% median improvement in the probability score for correct first next-token prediction.

% To conclude, by explicitly replacing the original word with a semantically equivalent but orthographically different atomization form, LLM can better reason and process character-level manipulation tasks more easily.

Through this systematic analysis of perturbation effects, we gain insights into how LLMs construct and maintain lexical representations throughout their processing pipeline. Notably, the strong performance of the fully atomized form particularly informs our method design in Section~\ref{sec:Method}, where we leverage this characteristic to improve character-level manipulation capabilities.

\section{Method}
\label{sec:Method}

Our analysis in Section \ref{sec:ProblemAnalysis} first identifies that LLMs struggle to effectively apply their intrinsic token knowledge to character-level reasoning. To address this, we propose the atomized word structure as a means to enhance LLMs' reasoning capabilities at the character level. Building on these insights, we introduce \method, a systematic approach that bridges token-level processing and character-level manipulation, enabling more precise and structured handling of character-level tasks.

\subsection{Task Formulation}

Character-level text manipulation serves as a fundamental building block in modern NLP systems. From data preprocessing to code generation and text normalization, these operations underpin numerous practical applications. 
While humans can perform such operations effortlessly, token-based LLMs encounter significant challenges due to their architectural constraints. In this work, we investigate three foundational character operations that capture core manipulation requirements while highlighting key technical challenges.

\begin{itemize}
    \item \textbf{Deletion} task requires removing specified characters while preserving word structure. Given a word $W$ and a target character $c_i \in W$, the task produces $W'$ such that $c_i \not \in W'$ while maintaining the order of remaining characters. For instance, removing 'l' from 'hello' should yeild 'heo'.
    \item \textbf{Insertion} task adds new characters at specific positions. Given a word $W$ and an anchor character $c_i \in W$, the task inserts $c_j$ after every occurance of $c_i \in W$ while keeping the rest of characters unchanged. When inserting 'a' after 'e' in 'hello', the output should be 'heallo'.
    \item \textbf{Substitution} task globally replaces characters throughout a word. Given a word $W$ and an target character $c_i \in W$, the substitution task is to replace each character $c_i$ with a new character $c_j$. For example, substituting 'l' with 'j' in 'hello' should produce 'hejjo'.
\end{itemize}

\subsection{\method}

\begin{figure}[!t]
\centering
\includegraphics[width=1.0\linewidth]{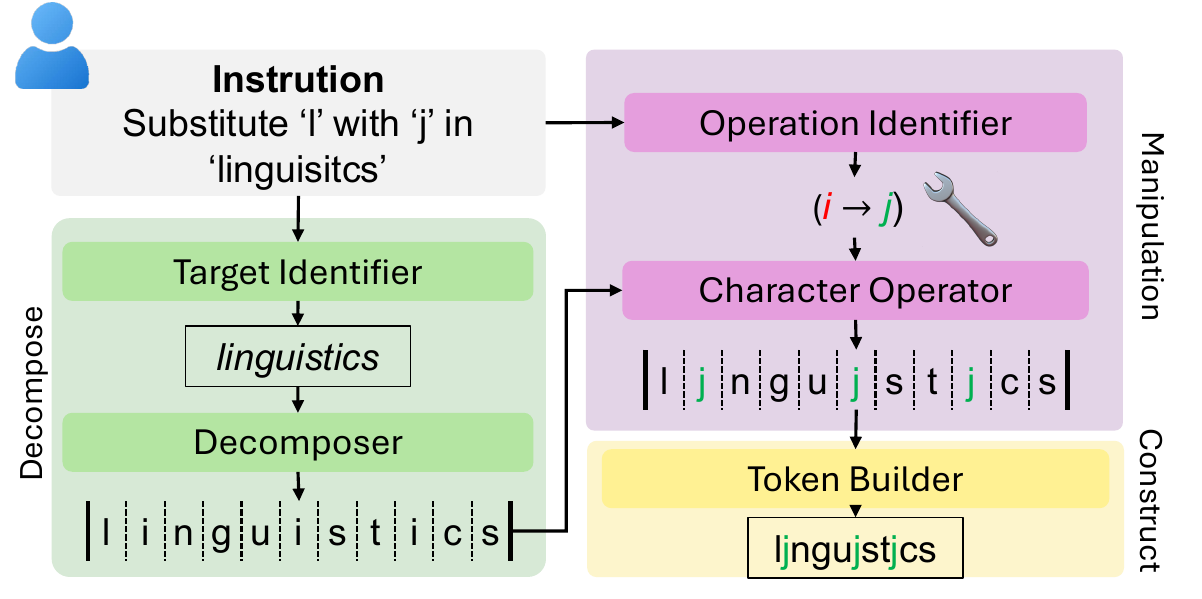}
\caption{Overview of our proposed method. We first decompose the token into an atomized character sequence. Then, we perform character-wise manipulations on the individual characters. Finally, we reconstruct the token from the modified sequence.}
\label{fig:method}
\end{figure}

Our key insight stems from Section \ref{sec:ProblemAnalysis}: while LLMs exhibit high spelling accuracy (97.4\%), they struggle with character-level reasoning, particularly in tasks involving letter retrieval and modification. Our analysis reveals that atomized words enhance LLMs' ability to reason about and manipulate individual characters effectively. Based on these findings, we propose a three-stage framework following the divide-and-conquer methodology, as illustrated in Figure \ref{fig:method}.

\paragraph{Stage I} (Token Atomization) bridges the character-level reasoning gap by transforming words into explicitly separated character sequences. Our findings indicate that LLMs internally encode character composition information but do not always utilize it effectively during reasoning tasks. By introducing controlled perturbations that separate adjacent characters, we activate this latent knowledge, enabling more precise character manipulation. The atomization step ensures that each character is independently accessible, mitigating tokenization artifacts that obscure intra-word structure.

\paragraph{Stage II} (Character-wise Manipulation) leverages the structured representation from Stage I to facilitate accurate and consistent character modifications. Since the model struggles with direct character retrieval yet excels at spelling, performing operations at the character level rather than within tokenized units significantly improves accuracy. This stage transforms complex word-level modifications into a sequence of simple, well-defined character operations, minimizing position-dependent errors and enhancing task reliability.

\paragraph{Stage III} (Token Construction) addresses the challenge of auto-correction by guiding the model through a controlled synthesis process. Instead of generating the modified word in a single step—where unwanted corrections may arise—we incrementally reconstruct the word from its modified character sequence. Our experimental results show that this approach prevents LLMs from reverting to common word forms, ensuring faithful execution of intended modifications.

This three-stage framework systematically aligns LLMs' demonstrated strengths with the requirements of character-level manipulation tasks. By leveraging atomized word structures, we unlock more effective character reasoning capabilities while maintaining computational efficiency. Figure \ref{fig:1} provides an example demonstrating how our method successfully executes a substitution task where direct prompting fails.

The following section presents comprehensive experiments demonstrating the effectiveness of our approach across different models, tasks, and word characteristics.

\section{Experiments}

In this section, we first introduce our implementation details and evaluation metrics, and then present various experiment results.

\paragraph{Implementation details} We use OpenAI official API for GPTs series evaluation. We set temperate to 0 and top\_p to 0.95 for all API requests. For system message and user message, please see Appendix \ref{apdix:prompts} for more details. 

\paragraph{Dataset construction} We select top 1K most frequently used English words \footnote{https://www.kaggle.com/datasets/rtatman/english-word-frequency/data} as input string to be manipulated. For \deltask and \subtask task, we randomly select one character within the word as the target to be removed or substituted with another different character. For \instask task, we first randomly select an existing character within the string as anchor and then randomly select another new character from the alphabet to insert after the previous anchor.

\paragraph{Evaluation metrics} Due to the deterministic natural of our tasks, we adopt exact match (EM) to evaluate the LLM's output is valid or not and the total accuracy is defined as: 
$$Acc = \frac{1}{N} \sum^{N}\text{EM}(y, \hat{y})$$
where N is the total number of testing samples and $y$ and $\hat{y}$ are model prediction and ground truth.

\subsection{Comparison Experiments}
Our experiments are conducted with various popular small-scale LLMs as well as proprietary commercial LLMs using different prompting strategies. 

The experimental results (see table \ref{tab:main_comparison_result}) demonstrate several key findings regarding the performance of different LLMs on character-level manipulation tasks.

\begin{table}[htbp]
\centering
\caption{Comparison of LLM Experiments Across Character-level Operations Tasks. \textbf{Del}, \textbf{Ins} and \textbf{Sub} are abbreviations for \deltask, \instask and \subtask tasks.}
\label{tab:main_comparison_result}
    \renewcommand{\arraystretch}{0.95} % vertical stretch
    \setlength{\tabcolsep}{5pt} % inter-col space
    \resizebox{0.9\linewidth}{!}{
    \begin{tabular}{lccc}
    \toprule
    \textbf{Model}& \textbf{Del} & \textbf{Ins} & \textbf{Sub} \\ 
    \midrule\midrule
    \gptthreepointfive w/ FS-1 & 0.875	& 0.159 & 0.663\\
    \gptthreepointfive w/ FS-4 & 0.903 & 0.162 & 0.439\\
    \gptthreepointfive w/ COT & 0.850 & 0.050 & 0.506 \\
    \gptthreepointfive w/ ours & \textbf{0.948} & \textbf{0.898} & \textbf{0.937}\\ \midrule
    \gptfouromini w/ FS-1 & 0.839 & 0.382 & 0.662\\
    \gptfouromini w/ FS-4 & 0.864 & 0.404 & 0.651\\
    \gptfouromini w/ COT & 0.881 & 0.462 & 0.788 \\
    \gptfouromini w/ ours & \textbf{0.918}  & \textbf{0.813}  & \textbf{0.916} \\ \midrule
    \llamathree w/ FS-1 & 0.469 & 0.070 & 0.145\\
    \llamathree w/ FS-4 & 0.572 & 0.071 & 0.331\\
    \llamathree w/ COT & 0.504 & 0.124 & 0.310 \\
    \llamathree w/ ours & \textbf{0.722} & \textbf{0.638} &\textbf{0.732}\\ \midrule
    \gemmatwo w/ FS-1 & 0.463 & 0.061 & 0.293\\
    \gemmatwo w/ FS-4 & 0.487 & 0.110 & 0.295\\
    \gemmatwo w/ COT & 0.584 & 0.102 & 0.444 \\
    \gemmatwo w/ ours & \textbf{0.769} & \textbf{0.510} & \textbf{0.694}\\
    \bottomrule
    \end{tabular}
    }
\end{table}

\paragraph{Overall Performance} Our proposed method consistently outperforms both few-shot~\cite{language2020} and chain-of-thought (COT)~\cite{cot2022} baselines across all models and tasks. Among all tested models, openAI's GPT-3.5 achieves the best performance with our method. The improvement is particularly significant for more complex operations like character insertion. 

\paragraph{Methodological Comparison} We considered different kinds of commonly used prompting strategy, including one/few-shot and COT. 
% Among different tasks, \ours archived x increase for \deltask, y increase for \instask, and z increase for \subtask with respect to other best competitors. 
Our method's consistent superior performance indicates that it addresses the limitations of both baseline approaches in handling character-level operations.

\paragraph{Task-specific Analysis} The experimental results reveal distinct patterns in different character manipulation tasks: \deltask task shows the highest baseline performance across all models, suggesting it might be the most intuitive operation for LLMs. \instask task: This proves to be the most challenging task for baseline methods, with few-shot and COT approaches struggling to achieve satisfactory results (average accuracy below 0.2 for some models). \subtask task: Performance on this task falls between deletion and insertion in terms of difficulty.

\paragraph{Model Architecture} Our experiment includes OpenAI`s GPT-3.5, GPT-4o-mini, Meta`s \llamathree~\cite{llama3}, and Google`s \gemmatwo~\cite{gemma2} for comparison. Although the results show that proprietary models outperform open-source models in all tasks, our method is LM-independent. In all the LLMs tested in the experiments, our method consistently outperforms other baseline methods.

\begin{figure}[!t]
\centering

\includegraphics[width=0.9\linewidth]{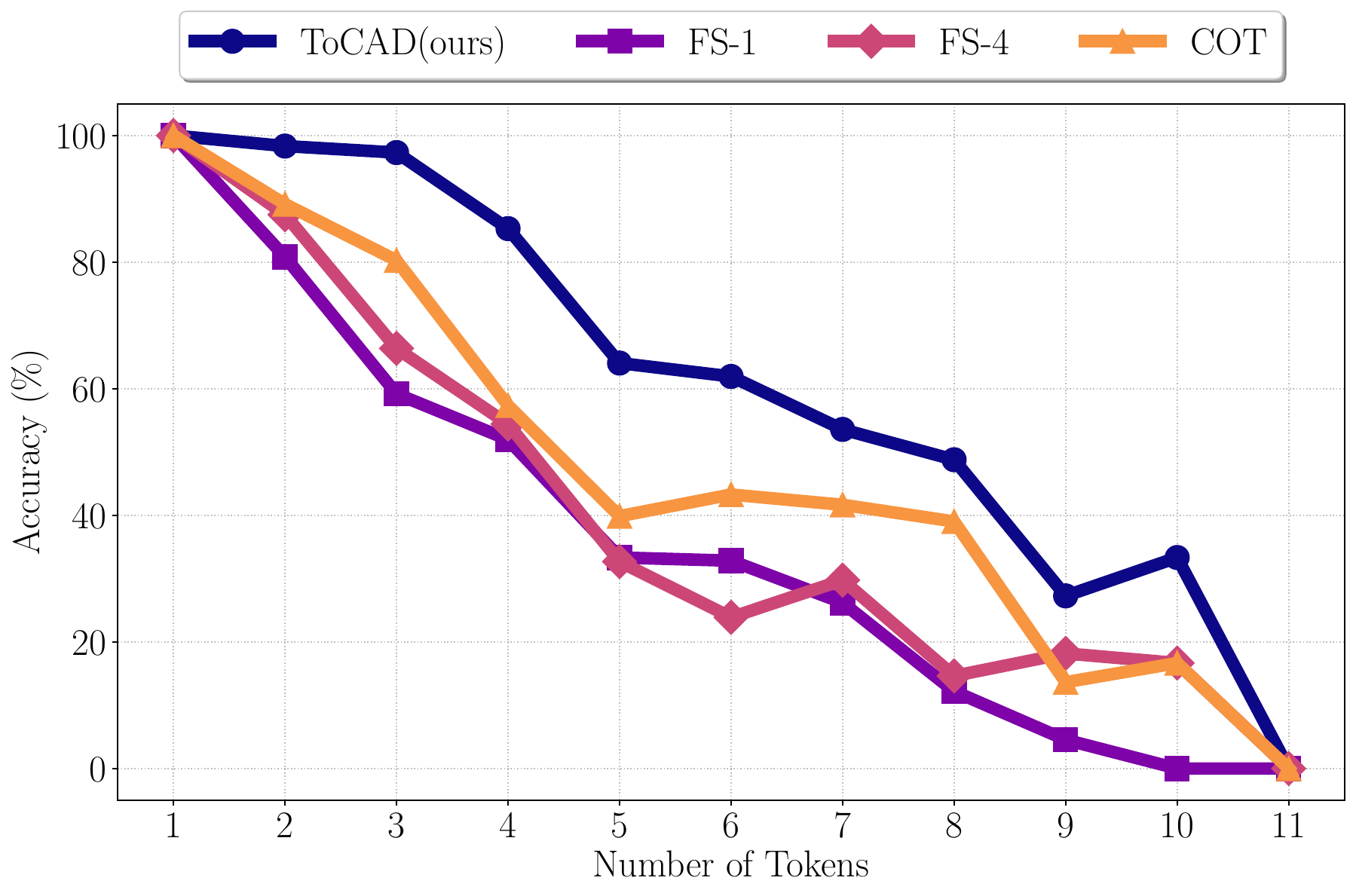}

\caption{\deltask accuracy w.r.t. word lengths on \gemmatwo. Our proposed method constantly outperforms other baselines. For more tasks and LMs comparison, check appendix~\ref{apdix:word_length}.}
\label{fig:ratio_vs_length}
\end{figure}

\paragraph{String Length} In the experiments, we also observed that the performance of LLM on string manipulation tasks is largely influenced by the length of the word, as illustrated in Figure \ref{fig:ratio_vs_length}. Specifically, model accuracy tends to decrease as string length increases. Although our approach is similarly affected by this pattern, it demonstrates a slower rate of performance deterioration compared to other baseline models, suggesting superior robustness in handling more challenging cases.

\subsection{Ablation Study}

\paragraph{Instruction-Following}

To evaluate the robustness of our method under varying parameter settings, we conducted experiments by modifying the instruction paradigm from zero-shot to few-shot. Specifically, we assessed the impact of different numbers of shots on the accuracy of each stage within our framework.

\begin{table}[htbp]
    \centering
    \renewcommand{\arraystretch}{1} % vertical stretch
    \setlength{\tabcolsep}{5pt} % inter-col space
    \resizebox{0.8\linewidth}{!}{
    \begin{tabular}{cllll}
    \toprule
    Stage & 0-shot & 1-shot & 2-shot & 3-shot \\ \midrule \midrule
    I & 0.993 & 0.997 & 0.997 & 0.999 \\
    II & 0.896 & 0.927 & 0.937 & 0.892 \\
    III & 0.636 & 0.673	& 0.685	& 0.671\\ \bottomrule
    \end{tabular}
    }
    
\caption{Performance at each stage for different instructions with varying numbers of examples.}
\label{table:ablation}
\end{table}

According to the ablation results (see table \ref{table:ablation}), increasing the number of examples in the few-shot setting did not lead to significant performance improvements across these three stages. This indicates that our method is not sensitive to prompt configuration and serves as a relatively stable and general framework for string manipulation.

\subsection{Case Study}

To demonstrate the effectiveness of the proposed method, we qualitatively analyzed failure cases from the evaluation dataset.

Two common error types were identified:

\paragraph{Error Type I: Auto-Correction} When the correct answer closely resembles the input word, LLMs tend to apply an internal correction mechanism. Instead of producing the expected output, they generate a semantically meaningful word. Examples are shown in the first row of the Table~\ref{tab:case_studies}.

\paragraph{Error Type II: Multi-Targets} Some tasks require modifying multiple occurrences of a character in a word. LLMs often stop processing after handling the first occurrence, leading to incomplete results. Examples are shown in the second row of the Table~\ref{tab:case_studies}.

\begin{table}[!tb]
\centering
    \renewcommand{\arraystretch}{1.0} % vertical stretch
    \setlength{\tabcolsep}{8pt} % inter-col space
    \resizebox{0.80\linewidth}{!}{

    \begin{tabular}{l c c c}
    \toprule
    \textbf{Type} & \textbf{Input} & \textbf{Expt} & \textbf{Pred} \\ \midrule \midrule
    \multirow{3}{*}{Type I} & movies & moviesq & movies \\ \cmidrule{2-4} 
     & include & iclude & include \\ \cmidrule{2-4} 
     & chat & chac & chat \\ \midrule
    \multirow{2}{*}{Type II} & whxich & whxichx & whxich \\ \cmidrule{2-4} 
     & data & dxtx & dxta \\ \bottomrule
    \end{tabular}
    }

\caption{Different error types with some failure cases as examples. \textbf{Input} and \textbf{Expt} are input string and expected ground truth, while \textbf{Pred} is the wrong output.}
\label{tab:case_studies}
\end{table}

\section{Conclusion}

In this article, we concentrated on improving LLMs' token-level understanding, with a specific focus on a series of fundamental character-level manipulation tasks. Our analysis reveals that LLMs face challenges in applying intrinsic token knowledge to character-level reasoning tasks. To address this, we introduced the atomized word structure, which elicits LLMs' ability to reason about token-level structural information.  Building on these insights, we propose a new approach, \method, to improve LLM character-level problem-solving skills. Based on extensive experiments, our method is significantly effective on all tasks with different LLMs. We believe our study has showcased a possible method to alleviate current LLMs' existing limitation on character-level understanding, inspiring future research concerning token understanding.

\section{Limitations \& Discussions}

This paper primarily focuses on publicly accessible instruction-finetuned token-level models. While character-level models may excel in tasks requiring character-level understanding, for now we believe improving current state-of-the-art LLMs is a more practical and cost-effective approach. Additionally, the Program-of-Thought~\cite{program2023} method, which generates code for string manipulation, does not address the fundamental issue of token-level understanding in LLMs. Recent inference-time computing models like OpenAI's o1 offer some potential for improving character-level manipulation, yet their efficiency in time and token usage is remain to be further explored.

% Entries for the entire Anthology, followed by custom entries
\bibliography{anthology,custom}
\bibliographystyle{acl_natbib}

\appendix

\section{prompts}
\label{apdix:prompts}

To compare our proposed method with frequently used baseline methods, we deployed single-shot prompt template (Figure.~\ref{fig:1shot-template}), 4-shot prompt template (Figure.~\ref{fig:4shot-template}) and Chain-of-Thought template (Figure.~\ref{fig:cot-template}) in our comparison experiment. 

\begin{figure}[htbp]
    \centering
    
\begin{tcolorbox}[width=\linewidth, fonttitle = \small\bfseries, title=One-Shot Prompt Template,colframe=gray!2!black,colback=gray!2!white,boxrule=1pt,boxsep=0pt,left=5pt,right=5pt,fontupper=\footnotesize, halign title = flush center]
\textbf{Deletion:} Delete every instance of a specified letter in a given word, based on the following examples:\textbackslash n\textbackslash ne.g.: Delete every instance of "a" in "alphabet". Answer: "lphbet"\textbackslash n\textbackslash nQuestion: Delete every instance of "\{\}" in "\{\}".
\tcbline

\textbf{Insertion:} Add the specified letter after every instance of the second specified letter in a given word, based on the following examples:\textbackslash n\textbackslash ne.g.: Add an "e" after every "a" in "alphabet". Answer: "aelphaebet"\textbackslash n\textbackslash nQuestion: Add an "\{\}" after every "\{\}" in "\{\}".

\tcbline

\textbf{Substitution:} Substitute the first specified letter with the second specified letter in a given word, based on the following examples:\textbackslash n\textbackslash ne.g.: Substitute "a" with "b" in "alphabet". Answer: "blphbbet"\textbackslash n\textbackslash nQuestion: Substitute "\{\}" with "\{\}" in "\{\}".

\end{tcolorbox}
\caption{Few-shot (n=1) prompt template.}
\label{fig:1shot-template}
\end{figure}

\begin{figure}[htbp]
    \centering
    
\begin{tcolorbox}[width=\linewidth, fonttitle = \small\bfseries, title=Three-Shot Prompt Template,colframe=gray!2!black,colback=gray!2!white,boxrule=1pt,boxsep=0pt,left=5pt,right=5pt,fontupper=\footnotesize, halign title = flush center]

\textbf{Deletion:} Delete every instance of a specified letter in a given word, based on the following examples:\textbackslash n\textbackslash n1. Delete every instance of "a" in "alphabet". Answer: "lphbet"\textbackslash n2. Delete every instance of "l" in "hello". Answer: "heo"\textbackslash n3. Delete every instance of "z" in "zebra". Answer: "ebra"\textbackslash n4. Delete every instance of "u" in "tongue". Answer: "tonge"\textbackslash n\textbackslash nQuestion: Delete every instance of "\{\}" in "\{\}".

\tcbline

\textbf{Insertion:} Add the specified letter after every instance of the second specified letter in a given word, based on the following examples:\textbackslash n\textbackslash n1. Add an "e" after every "a" in "alphabet". Answer: "aelphaebet"\textbackslash n2. Add an "l" after every "l" in "hello". Answer: "hellllo"\textbackslash n3. Add an "t" after every "z" in "zebra". Answer: "ztebra"\textbackslash n4. Add an "f" after every "u" in "tongue". Answer: "tongufe"\textbackslash n\textbackslash nQuestion: Add an "\{\}" after every "\{\}" in "\{\}".

\tcbline

\textbf{Substitution:} Substitute the first specified letter with the second specified letter in a given word, based on the following examples:\textbackslash n\textbackslash n1. Substitute "a" with "b" in "alphabet". Answer: "blphbbet"\textbackslash n2. Substitute "h" with "e" in "hello". Answer: "eello"\textbackslash n3. Substitute "z" with "a" in "zebra". Answer: "aebra"\textbackslash n4. Substitute "u" with "e" in "tongue". Answer: "tongee"\textbackslash n\textbackslash nQuestion: Substitute "\{\}" with "\{\}" in "\{\}".

\end{tcolorbox}
\caption{Few-shot(n=4) prompt template.}
\label{fig:4shot-template}
\end{figure}

\begin{figure}[htbp]
    \centering
    
\begin{tcolorbox}[width=\linewidth, fonttitle = \small\bfseries, title=Chain-of-Thought Prompt Template,colframe=gray!2!black,colback=gray!2!white,boxrule=1pt,boxsep=0pt,left=5pt,right=5pt,fontupper=\footnotesize, halign title = flush center]

\textbf{Deletion:} Delete every instance of "\{\}" in "\{\}". Show you reasoning process step by step. Please provide the final answer at the end with "Answer:".

\tcbline

\textbf{Insertion:} Add an "\{\}" after every "\{\}" in "\{\}". Show you reasoning process step by step. Please provide the final answer at the end with "Answer:".

\tcbline

\textbf{Substitution:} Substitute "\{\}" with "\{\}" in "\{\}". Show you reasoning process step by step. Please provide the final answer at the end with "Answer:".

\end{tcolorbox}
\caption{Chain-of-Thought (CoT) prompt template.}
\label{fig:cot-template}
\end{figure}

\section{Attentions}
\label{apdix:attentions}

This section of the appendix presents another example (see Figure~\ref{fig:appedix_ratio_vs_length}) illustrating how an atomized word structure can enhance and reinforce an LLM's structural reasoning ability at the character level. This observation motivated us to later develop a framework that enables LLMs to process character-level manipulation tasks more accurately.

\begin{figure*}[!tb]
    \centering
    \includegraphics[width=1\textwidth]{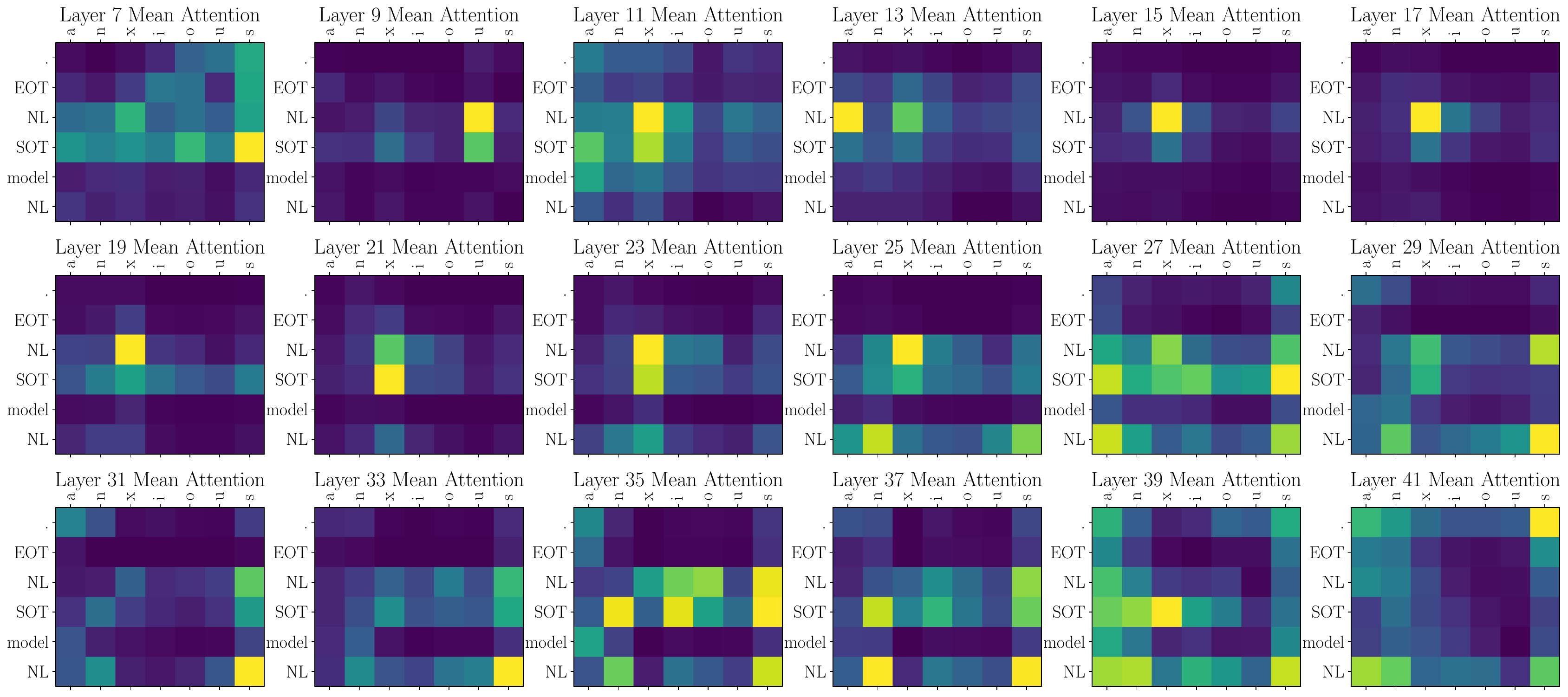}
    \caption{Averaged attention matrix across heads on different layers with a special focus on tokens of interest. The darker the color of the heat map cell represents smaller the attention value, and vice versa. 
    Notice that to maximize the use of the limited space, we renamed certain special tokens (<start\_of\_turn> $\rightarrow$ SOT, <end\_of\_turn> $\rightarrow$ EOT, \textbackslash n $\rightarrow$ NL) and cropped the attention matrices to include only the relevant tokens. In our case, the token "x" in the x-axis is the target letter to remove from the word "anxious". Meanwhile, "an" is the first expected output token.}
    \label{fig:attn_mat_anxious}
\end{figure*}

\section{Robustness}
\label{apdix:word_length}

\begin{figure*}[!tb]
 \centering
 % First column (Gemma2)
    \begin{subfigure}[b]{0.30\textwidth}
        \centering
        \includegraphics[width=\textwidth]{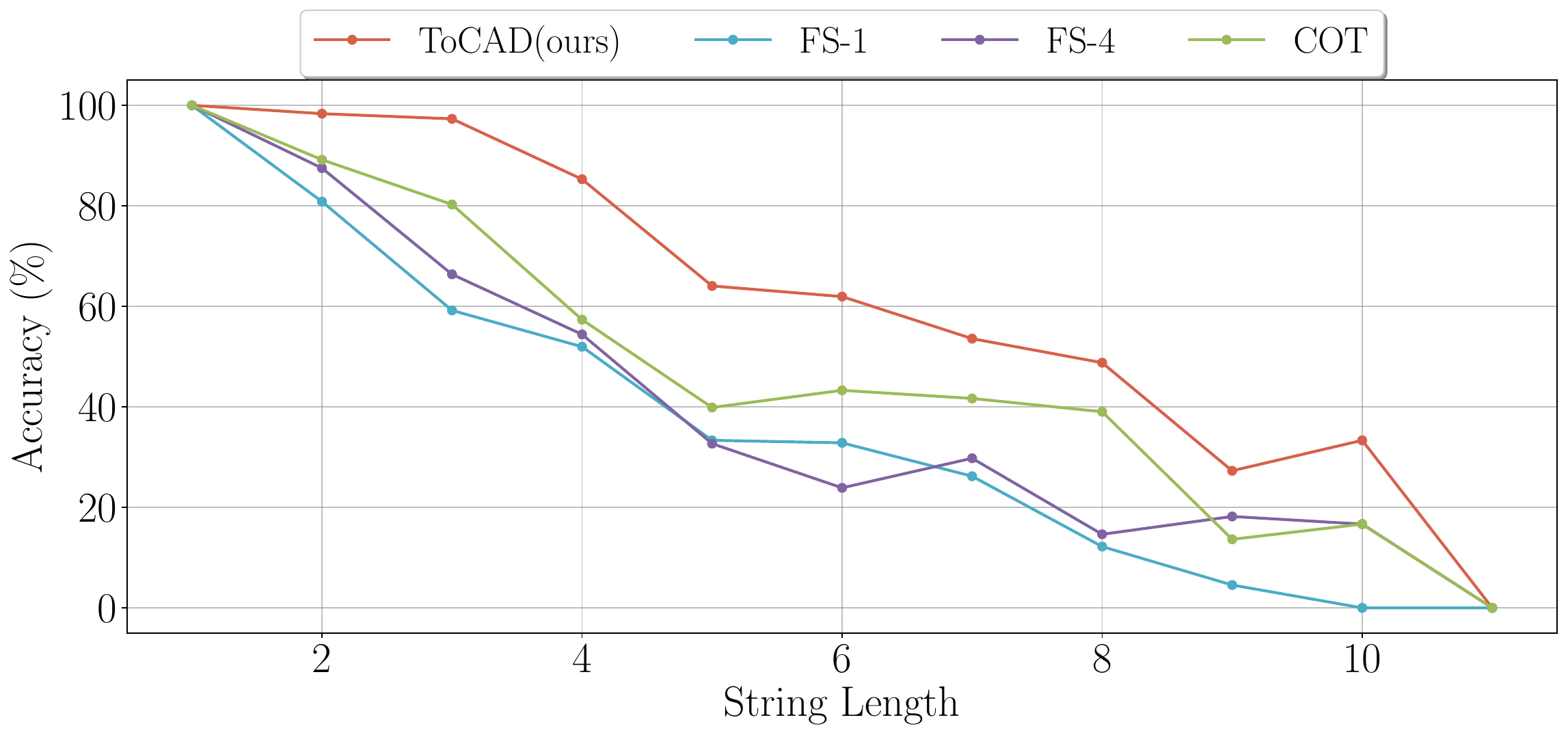}
        \caption{\textbf{Del} on \gemmatwo}
    \end{subfigure}
    \begin{subfigure}[b]{0.30\textwidth}
        \centering
        \includegraphics[width=\textwidth]{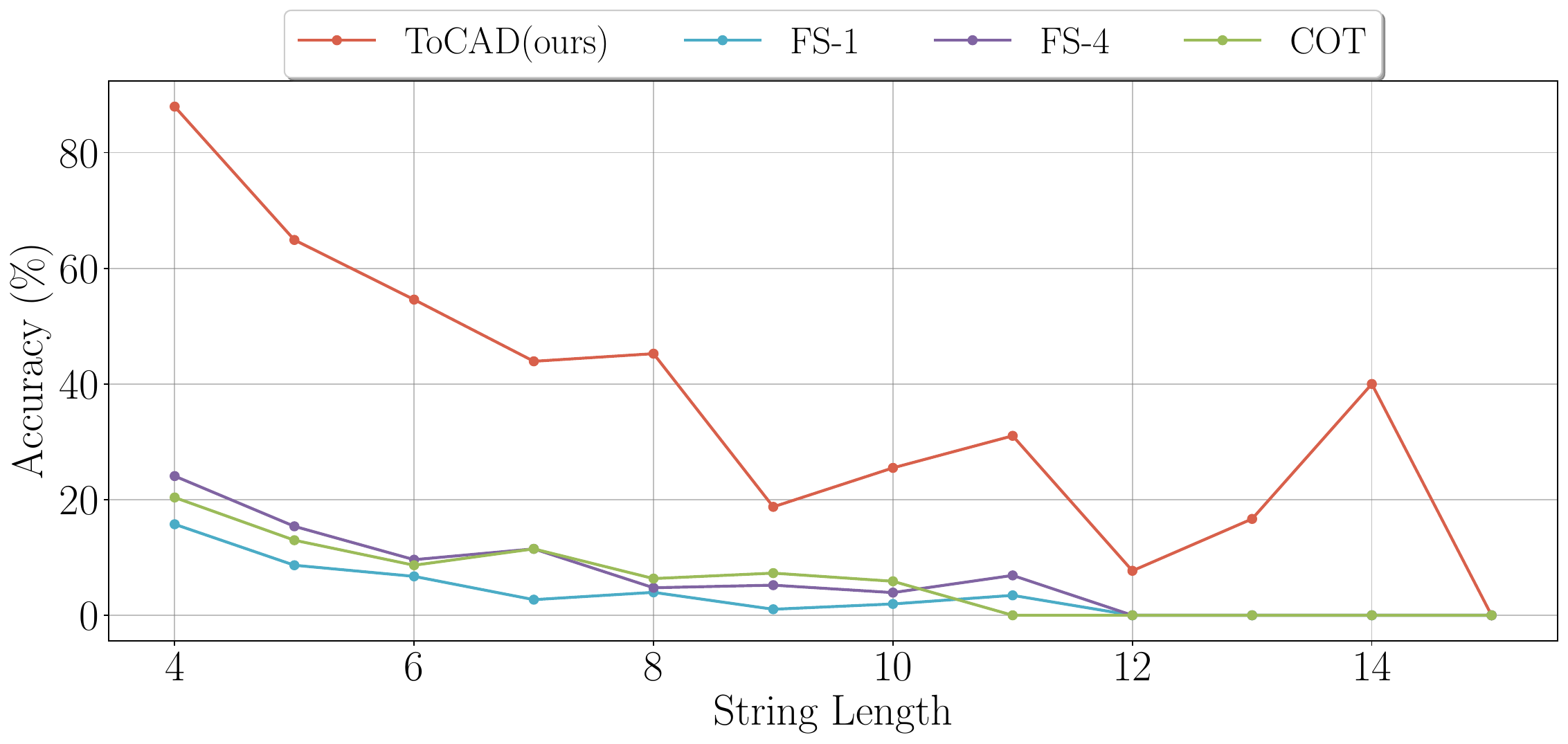}
        \caption{\textbf{Ins} on \gemmatwo}
    \end{subfigure}
    \begin{subfigure}[b]{0.30\textwidth}
        \centering
        \includegraphics[width=\textwidth]{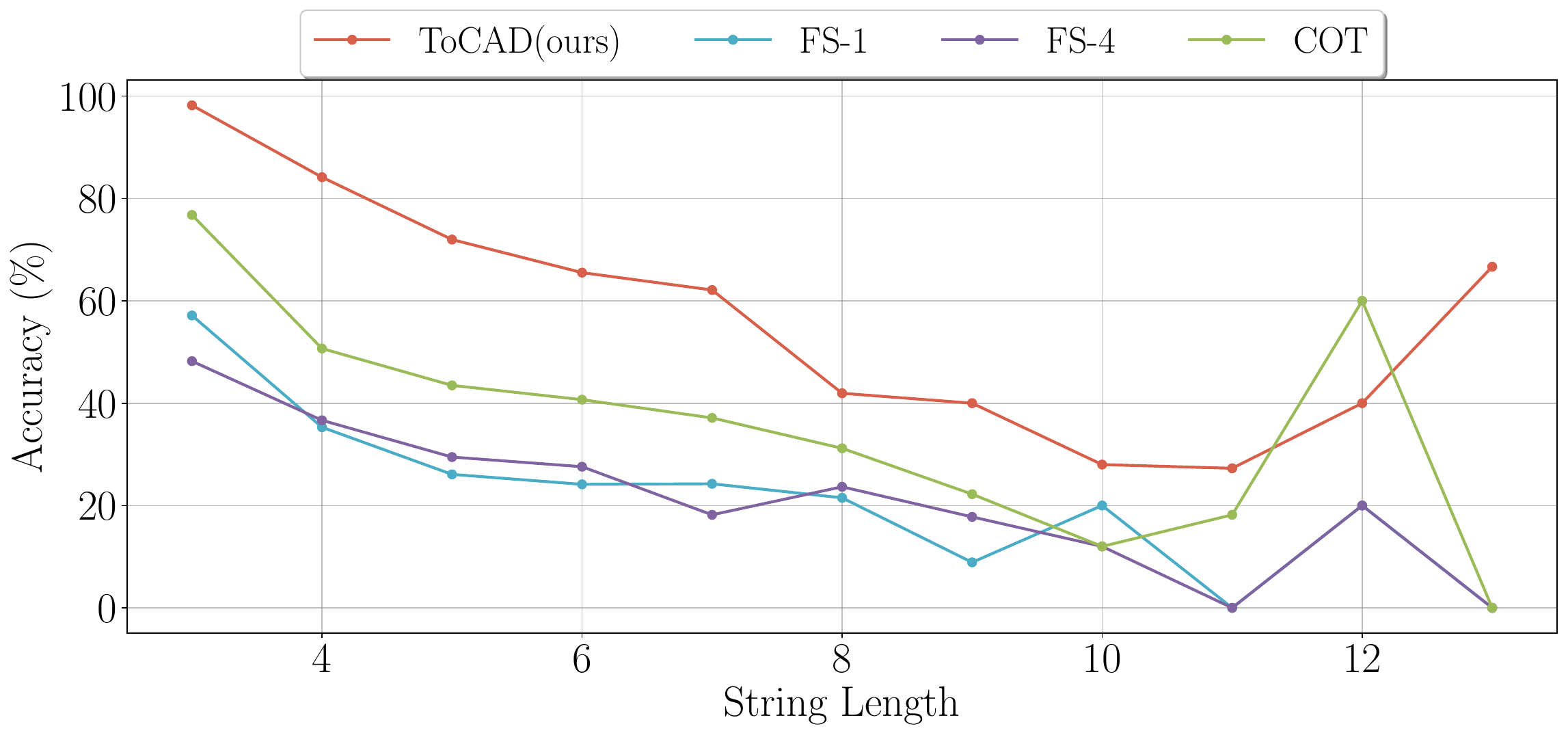}
        \caption{\textbf{Sub} on \gemmatwo}
    \end{subfigure}
    
    \vfill
    
    % Second column (LLaMA3)
    \begin{subfigure}[b]{0.30\textwidth}
        \centering
        \includegraphics[width=\textwidth]{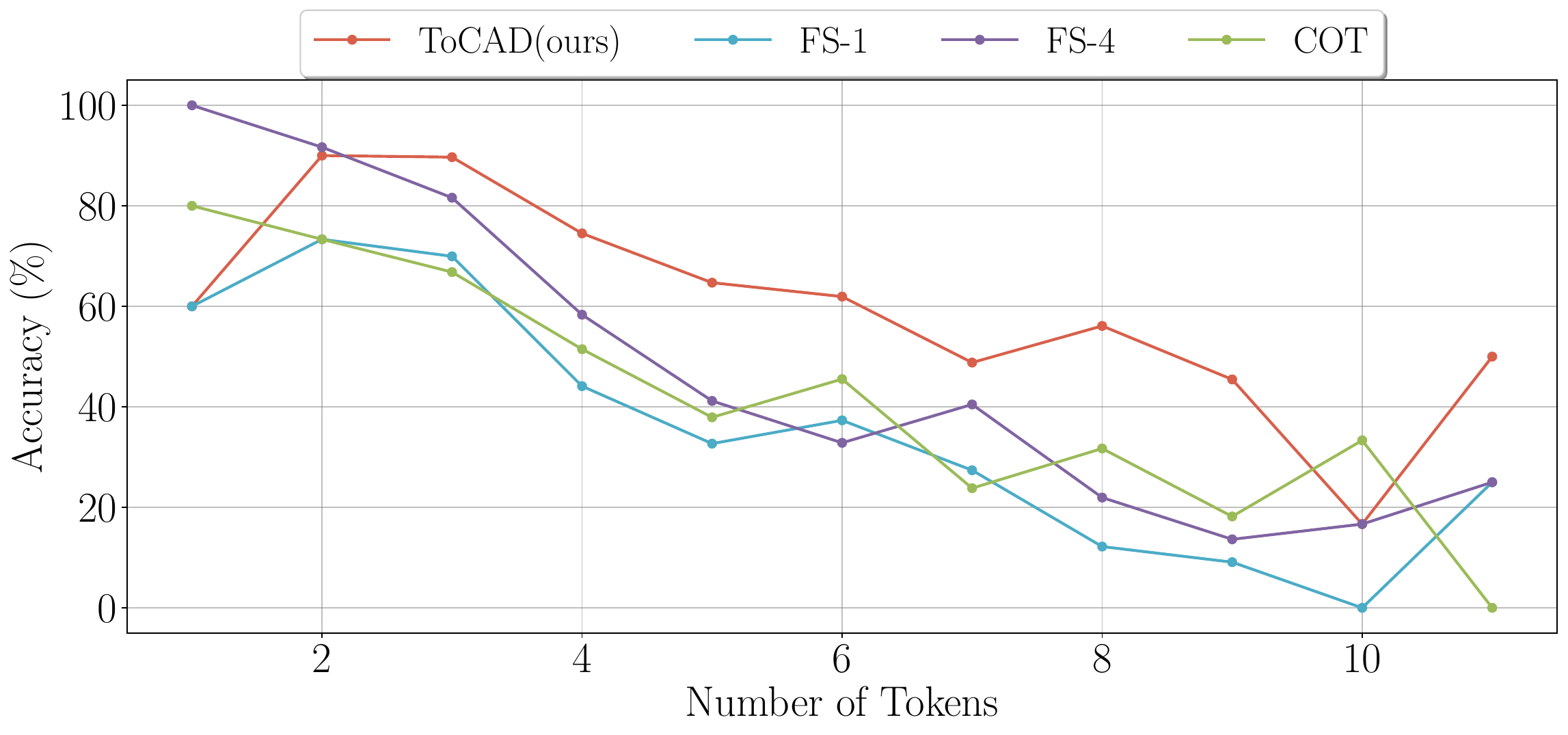}
        \caption{\textbf{Del} on \llamathree}
    \end{subfigure}
    \begin{subfigure}[b]{0.30\textwidth}
        \centering
        \includegraphics[width=\textwidth]{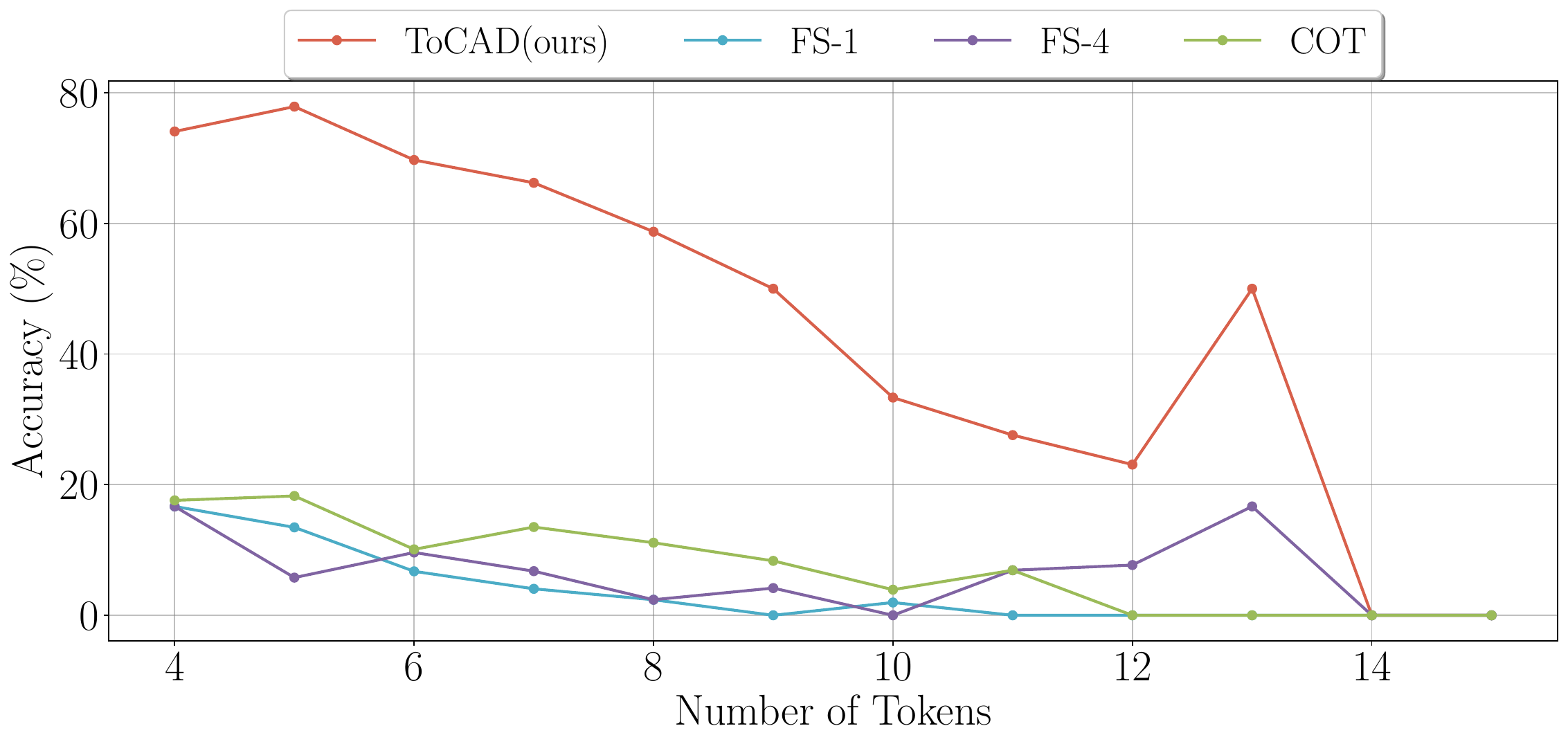}
        \caption{\textbf{Ins} on \llamathree}
    \end{subfigure}
    \begin{subfigure}[b]{0.30\textwidth}
        \centering
        \includegraphics[width=\textwidth]{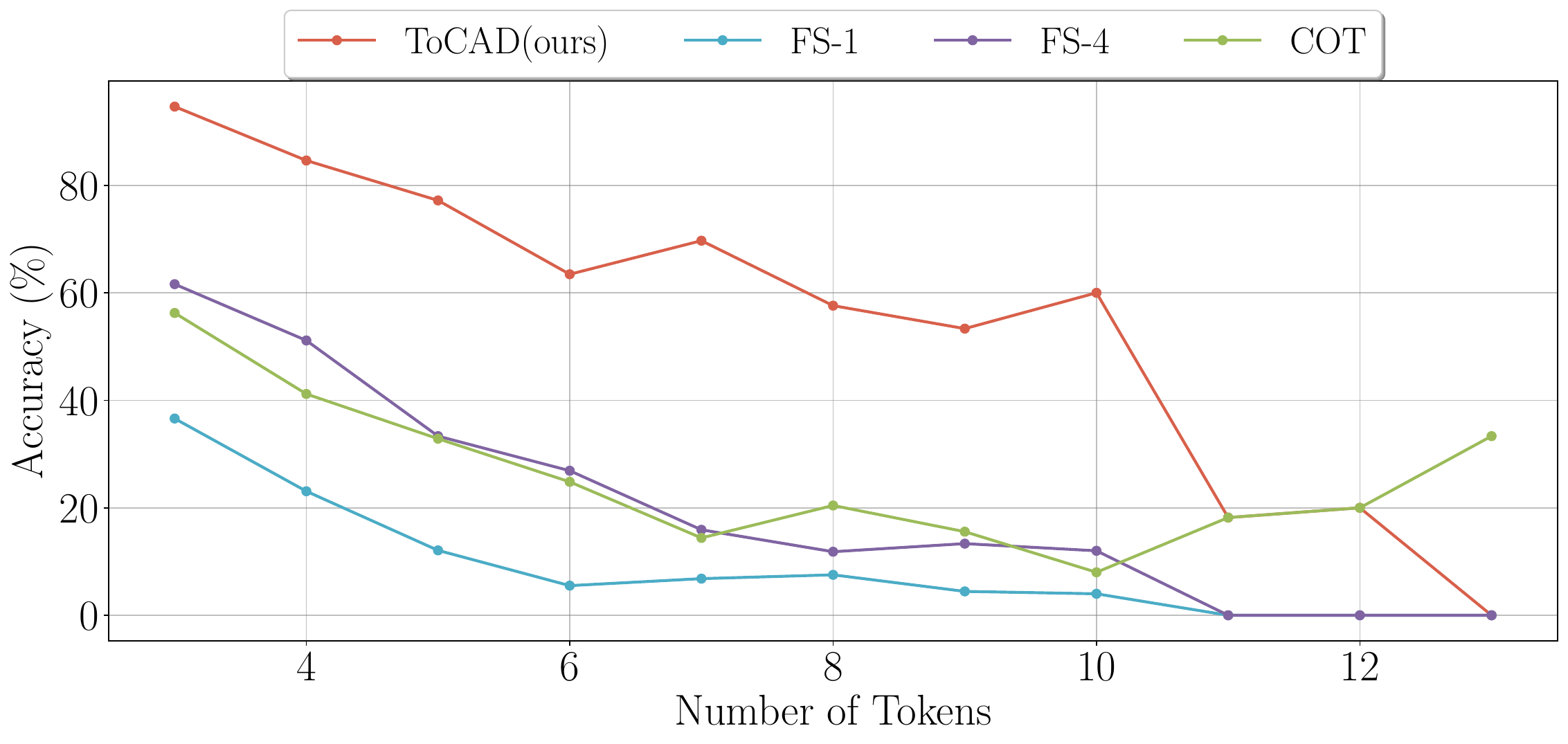}
        \caption{\textbf{Sub} on \llamathree}
    \end{subfigure}
    
    \vfill
    
    % Third column (GPT-3.5 Turbo)
    \begin{subfigure}[b]{0.30\textwidth}
        \centering
        \includegraphics[width=\textwidth]{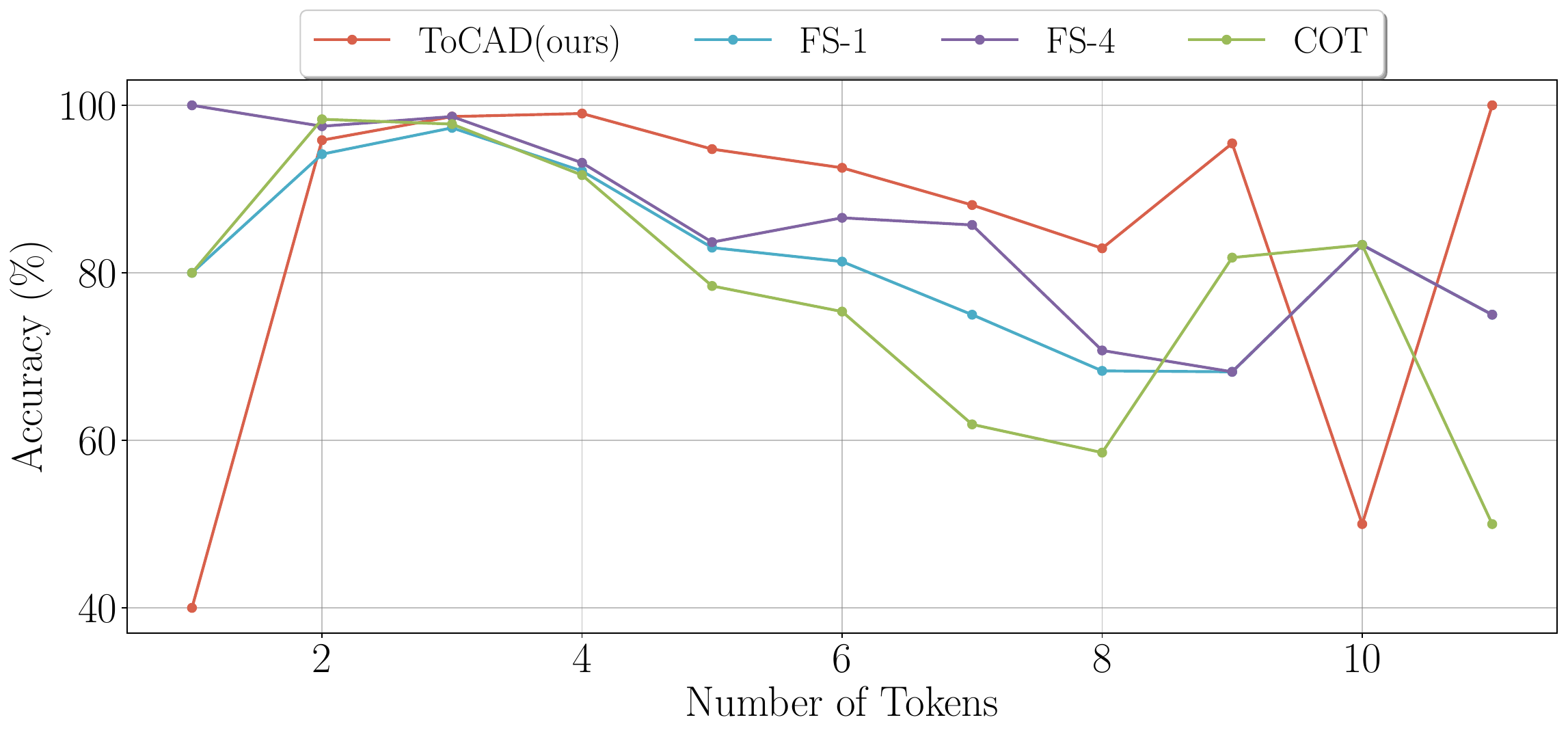}
        \caption{\textbf{Del} on \gptthreepointfive}
    \end{subfigure}
    \begin{subfigure}[b]{0.30\textwidth}
        \centering
        \includegraphics[width=\textwidth]{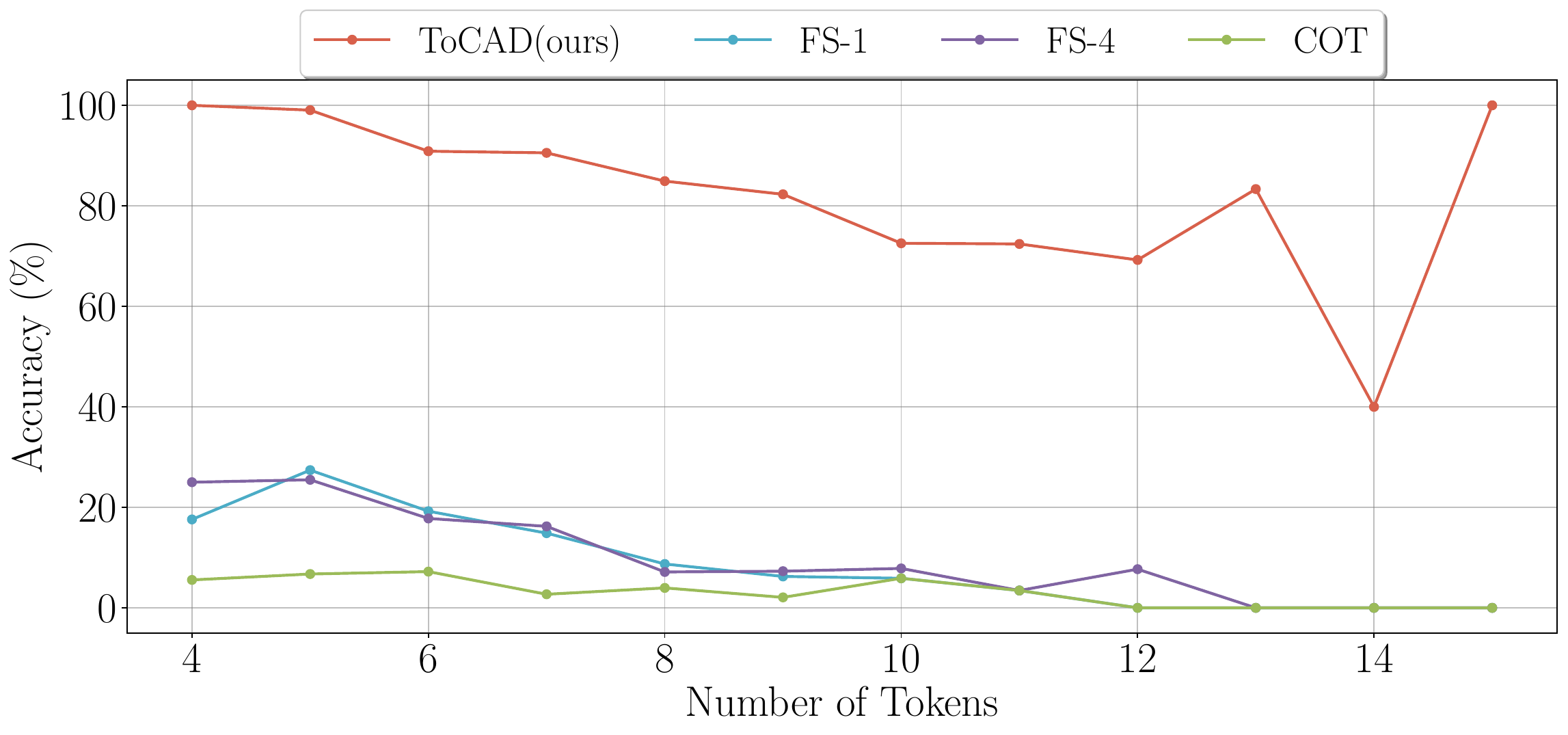}
        \caption{\textbf{Ins} on \gptthreepointfive}
    \end{subfigure}
    \begin{subfigure}[b]{0.30\textwidth}
        \centering
        \includegraphics[width=\textwidth]{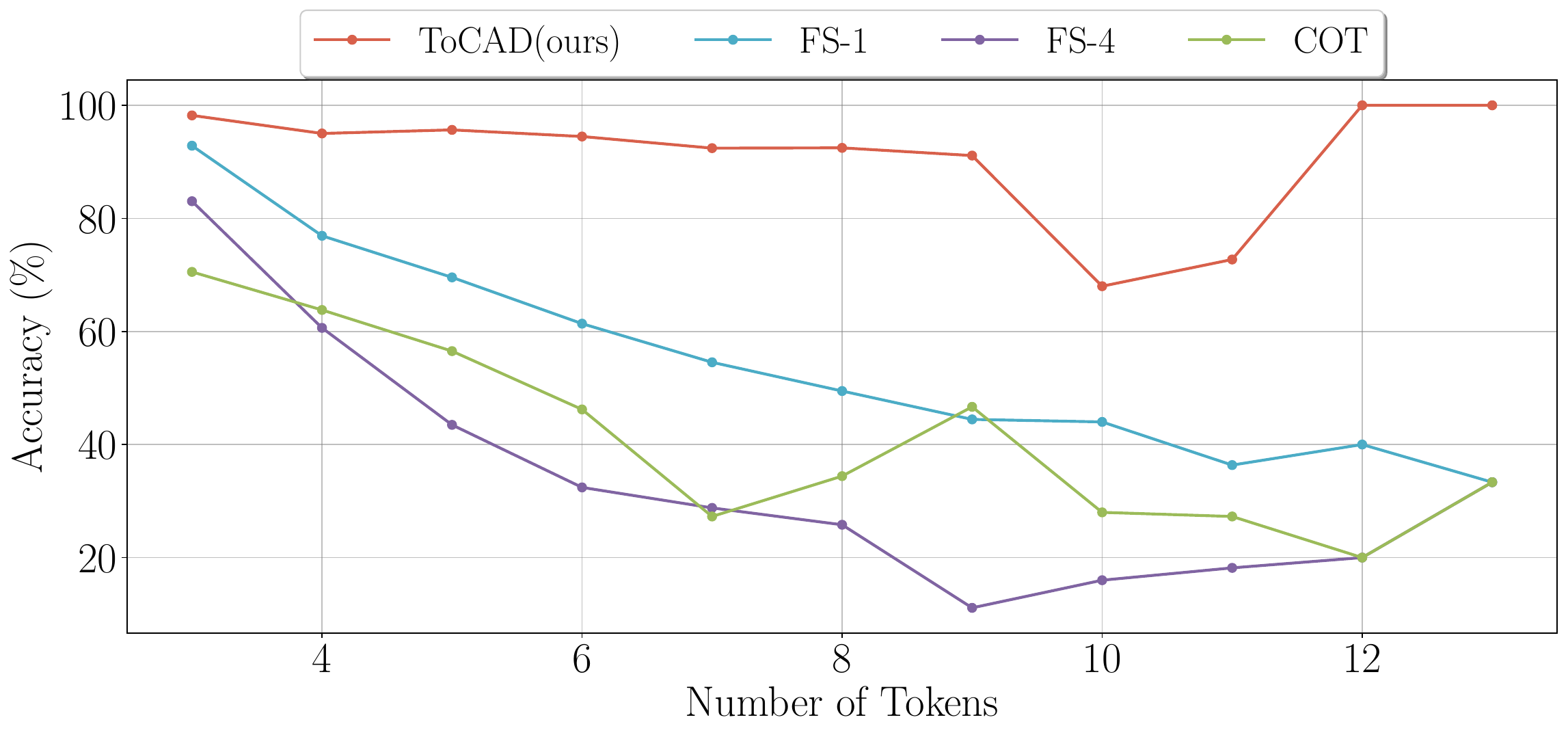}
        \caption{\textbf{Sub} on \gptthreepointfive}
    \end{subfigure}
    
    \vfill
    
    % Fourth column (GPT-4o Mini)
    \begin{subfigure}[b]{0.30\textwidth}
        \centering
        \includegraphics[width=\textwidth]{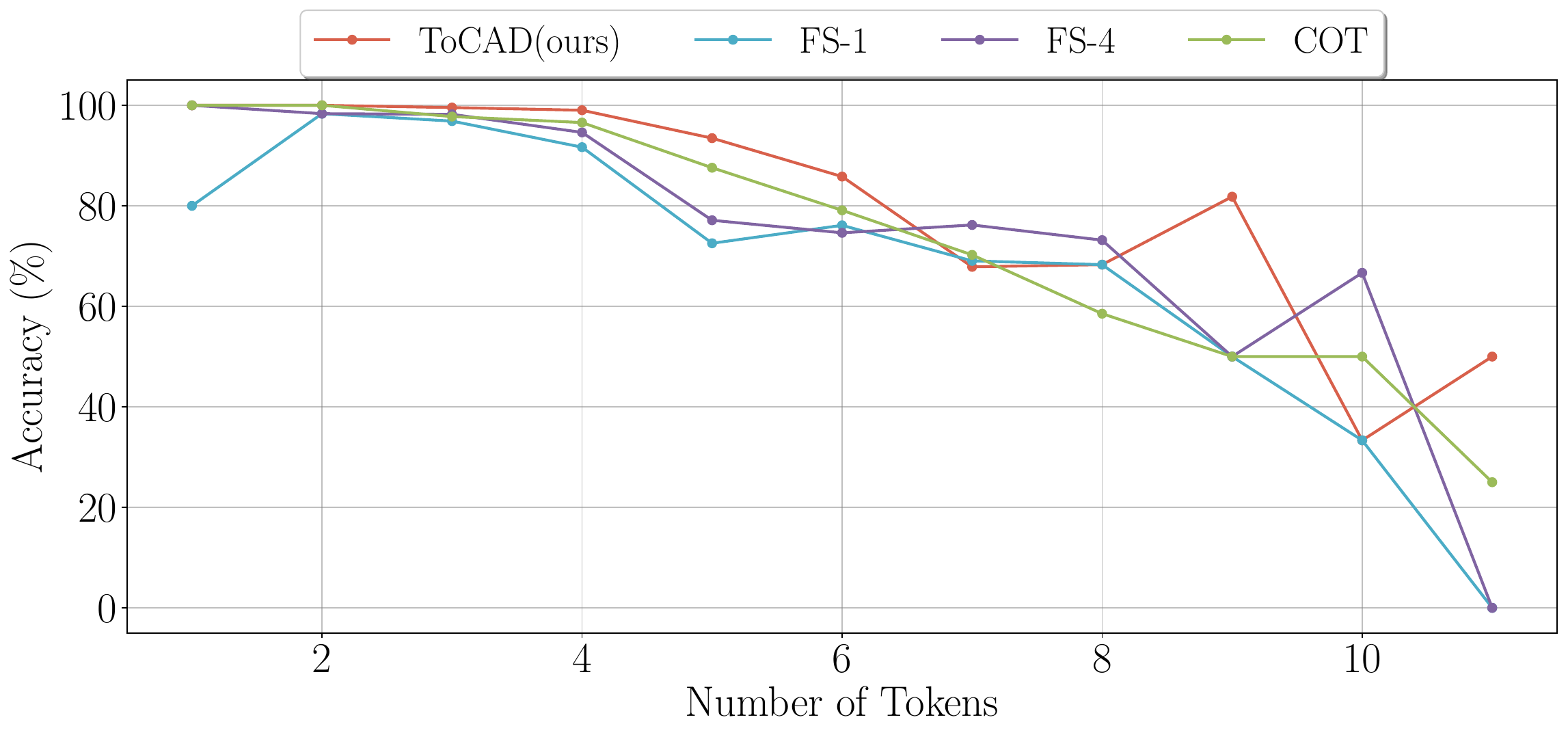}
        \caption{\textbf{Del} on \gptfouromini}
    \end{subfigure}
    \begin{subfigure}[b]{0.30\textwidth}
        \centering
        \includegraphics[width=\textwidth]{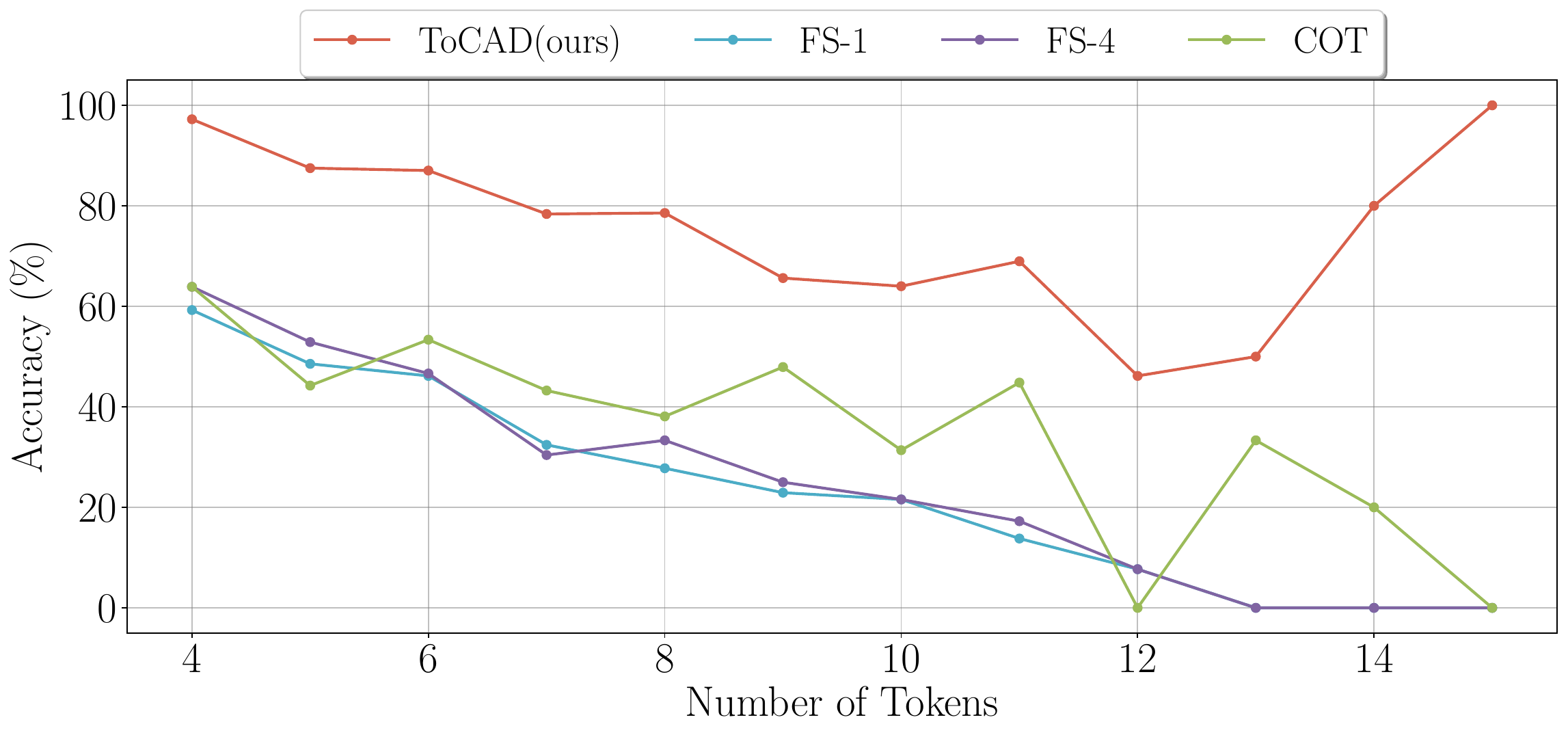}
        \caption{\textbf{Ins} on \gptfouromini}
    \end{subfigure}
    \begin{subfigure}[b]{0.30\textwidth}
        \centering
        \includegraphics[width=\textwidth]{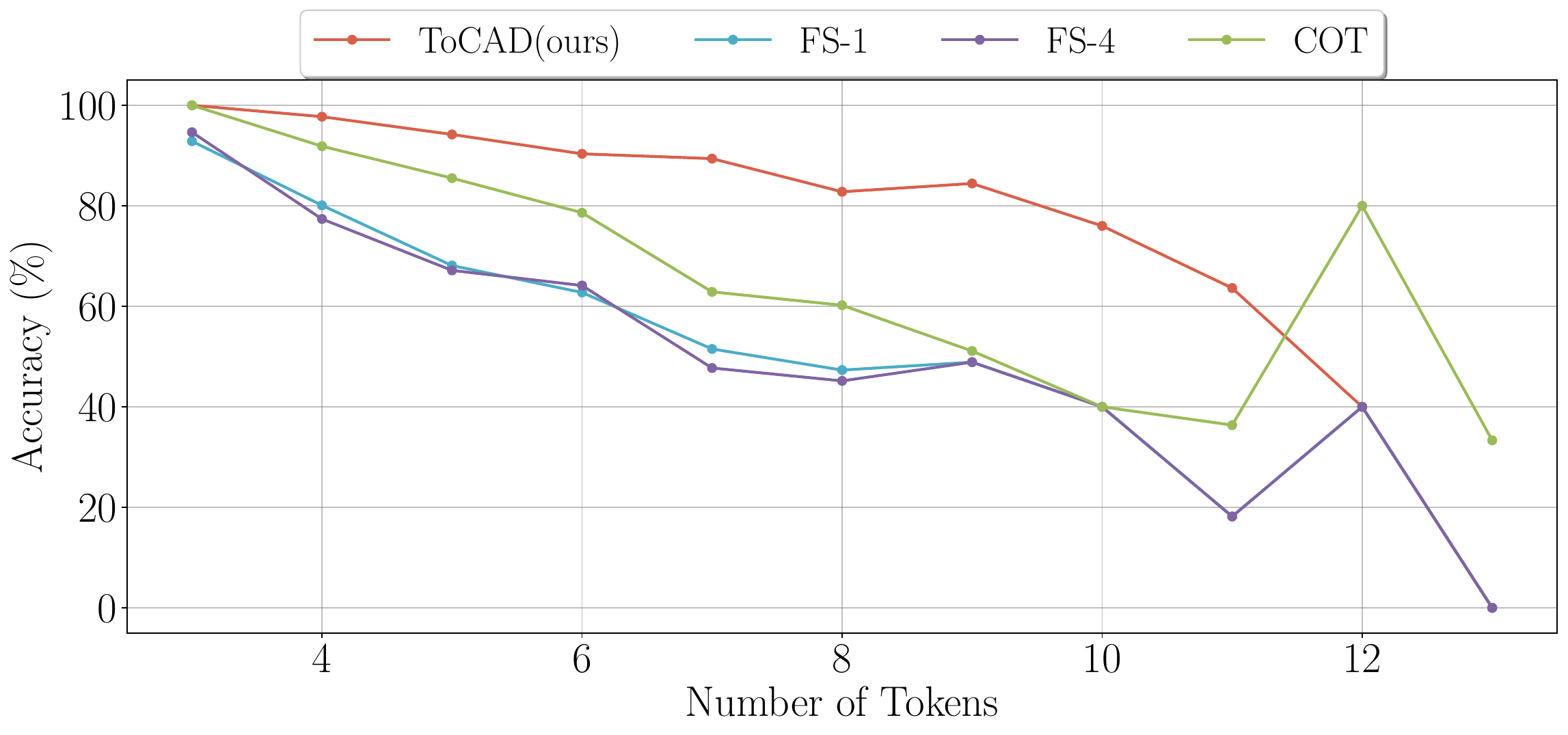}
        \caption{\textbf{Sub} on \gptfouromini}
    \end{subfigure}
 
\caption{Character-level manipulation accuracy comparison of different baseline methods on various string lengths across four different LLMs(\gemmatwo, \llamathree, \gptfouromini, and \gptthreepointfive). Our proposed method constantly outperforms other comparing baseline methods.}
\label{fig:appedix_ratio_vs_length}
\end{figure*}

This appendix section provides additional experimental results, comparing baseline methods with our proposed methods across varying word lengths, using different large language models (LLMs) on multiple tasks (Figure~\ref{fig:appedix_ratio_vs_length}). The results demonstrate that our methods consistently outperform the baseline methods across all scenarios.

\end{document}